%% file: Main.tex
\documentclass[journal,transmag]{IEEEtran}
\ifCLASSINFOpdf
\else
\fi
\usepackage{algpseudocode}
\usepackage{algorithm}
\usepackage{multirow}
\usepackage{array}
\usepackage[lofdepth,lotdepth]{subfig}
\usepackage{booktabs}
\usepackage{graphicx}
\usepackage{xcolor}
\usepackage{soul}
\usepackage[utf8]{inputenc}
\usepackage[figuresright]{rotating}
\usepackage{subfig}
\usepackage{float}
\usepackage{subfloat}
\usepackage{stfloats}
\usepackage{mdwmath}
\usepackage{amsmath, amsthm, amssymb}
\usepackage{cases}
\usepackage{tcolorbox}
\usepackage{authblk}
\usepackage[normalem]{ulem}
\usepackage{siunitx}
\usepackage{authblk}
\usepackage{hyperref}
\usepackage{tikz}
\usepackage{ulem}\usepackage{cancel}
\usepackage{listings}
\usepackage{wrapfig}
\usepackage{enumitem}                           
\usepackage{array}                              
\usepackage{makecell}                           
\usepackage{amssymb}                            
\usepackage{mathtools}
\usepackage{setspace}
\usepackage{multirow}
\usepackage{multicol}
\usepackage{hhline}
\usepackage{caption}                            
\usepackage{dirtytalk}                          
\usepackage{algpseudocode}                      
\usepackage{algorithm}                          
\usepackage{diagbox}                            
\usepackage{bm}

\usepackage{multicol}
\usepackage{soul}
\usepackage{booktabs}
\usepackage{colortbl}
\begin{document}
%
\title{A Personalized Zero-Shot ECG Arrhythmia Monitoring System: \\ From Sparse Representation Based Domain Adaption to \\ Energy Efficient Abnormal Beat Detection for Practical ECG Surveillance}

\author[1]{Mehmet Yama\c{c}$^*$}
\author[1]{Mert Duman$^*$}
\author[1]{Ilke Adalıoğlu}
\author[2]{Serkan Kiranyaz}
\author[1]{Moncef Gabbouj}
\affil[1]{Tampere University, Faculty of Information Technology and Communication Sciences, Tampere, Finland}
\affil[2]{Department of Electrical Engineering, Qatar University, Qatar}


\maketitle

\def\thefootnote{*}\footnotetext{These authors contributed equally to this work}
\begin{abstract}

\textit{Abstract}---This paper proposes a low-cost and highly accurate ECG-monitoring system intended for personalized early arrhythmia detection for wearable mobile sensors. Earlier supervised approaches for personalized ECG monitoring require both abnormal and normal heartbeats for the training of the dedicated classifier. However, in a real-world scenario where the personalized algorithm is embedded in a wearable device, such training data is not available for healthy people with no cardiac disorder history. 
In this study, (i) we propose a null space analysis on the healthy signal space obtained via sparse dictionary learning, and investigate how a simple null space projection or alternatively regularized least squares-based classification methods can reduce the computational complexity, without sacrificing the detection accuracy, when compared to sparse representation-based classification. (ii) Then we introduce a sparse representation-based domain adaptation technique in order to project other existing users' abnormal and normal signals onto the new user's signal space, enabling us to train the dedicated classifier without having any abnormal heartbeat of the new user. Therefore, zero-shot learning can be achieved without the need for synthetic abnormal heartbeat generation. An extensive set of experiments performed on the benchmark MIT-BIH ECG dataset shows that when this domain adaptation-based training data generator is used with a simple 1-D CNN classifier, the method outperforms the prior work by a significant margin. (iii) Then, by combining (i) and (ii), we propose an ensemble classifier that further improves the performance. This approach for zero-shot arrhythmia detection achieves an average accuracy level of 98.2\% and an F1-Score of 92.8\%. Finally, a personalized energy-efficient ECG monitoring scheme is proposed using the above-mentioned innovations. The software implementation of the proposed technique is publicly available at \url{https://github.com/MertDuman/Zero-Shot-ECG}.

\end{abstract}


\IEEEpeerreviewmaketitle

\section{Introduction}
\IEEEPARstart{A}{n} abnormal heart rhythm, i.e., cardiac arrhythmia can be a predictor of a major health threat such as stroke or heart failure. In fact, the past statistics show that cardiovascular diseases continue to be the number one cause of death, i.e, $32.84\%$ of all deaths in 2019 were caused by cardiovascular diseases \cite{murray2020global}. Real-time detection of cardiac arrhythmias using computer technology can significantly reduce the number of casualties. Numerous methods have been proposed for identifying electrocardiogram (ECG) waveforms using signal processing and machine learning, such as frequency analysis \cite{frequency}, support vector machines (SVMs) \cite{SVM}, statistical analysis \cite{statistical}, etc. The majority of these methods, however, are based on a set of handcrafted features, making them inflexible to changes in ECG patterns. Depending on the circumstances and even from heartbeat to heartbeat, ECG morphology can show variations \cite{hoekema2001geometrical}. Additionally, ECG morphology is unique to each individual according to their cardiovascular system \cite{ECGDataset1}. As a result, robust person identification from ECG signals has become a popular research topic recently \cite{ECGBiometrics1, ibtehaz2021edith}. Therefore, it is essential to design a person-based (personalized) arrhythmia detection system, which is not possible by the traditional methods mentioned above. 

A personalized classification and arrhythmia detection system is proposed in \cite{kiranyaz2016real, patient_spesific1, patient_spesific2}. In \cite{zhai2018automated}, Zhai \textit{et al.} represented 1-D ECG signals in 2-D by taking the outer product of two adjacent dual-beats. This 2-D coupling matrix captures the morphology of a single heartbeat as well as the temporal morphological characteristics of the adjacent beats. In \cite{li2019automated}, various features (e.g. R-R intervals, beat-to-beat correlation, etc.) are extracted from the ECG signals and stacked to create 3-D inputs to a convolutional neural network (CNN). 
Nevertheless, such methods require both normal and abnormal heartbeats from a patient in order to train a dedicated classifier. Therefore, they cannot serve as an early detection system for an otherwise healthy person who has no history of heart disease. As an alternative, the landmark study in \cite{kiranyaz2017personalized} has proposed a system to model the degradation between normal and abnormal beats by learning linear filters from several arrhythmia patients. After that, for a new user, the potential abnormal beats can be synthesized directly from the normal beats, and finally, a dedicated 1-D CNN is trained over synthesized abnormal heartbeats and real normal heartbeats to monitor signs of arrhythmia of the user.

Sparse approximation (SA) is a procedure that involves determining a set of waveforms known as a dictionary or a basis set. Then, a signal can be embodied as a linear combination of a small number of these predefined dictionary elements, or in other words, as a sparse code representation.
Some of the recent applications of sparse decomposition include compressed sensing (CS) \cite{CS}, classification \cite{yamacc2021convolutional}, and encrypted ECG monitoring \cite{yamacECG}. In recent works \cite{huang2012sparse, mathews2015leveraging} SA has been successfully used in ECG classification tasks. These methods, however, attempt to learn SA dictionaries from the extracted features, rather than directly from the raw data. Therefore, their performance is still limited by the discrimination power of the handcrafted features.

An alternative approach that bypasses the need to collect abnormal ECG data  \cite{carrera2016ecg}, and uses only normal heartbeats in the training phase relies on an outlier detection scheme. The methods in \cite{carrera2016ecg}, and \cite{Elad} propose a sparse approximation method that learns the dictionary directly from the normal beats of a specific user without the need for any abnormal beat synthesis. The underlying principle is that normal heartbeats live in the learned subspace while abnormal heartbeats do not. Thus, during the monitoring phase, a heartbeat is considered abnormal if it cannot be sparsely represented in the person-specific dictionary. However, they use only 1-D features and a simple classification technique to detect the anomaly. Adler \textit{et al.} \cite{Elad} have tried to separate sparse and non-sparse residuals by using consecutive heartbeats via group sparsity. As a feature, they use the energy of each sparse error. The methods proposed in \cite{carrera2016ecg}, and \cite{carrera2019online}, have also used a simple feature which is the energy of sparse approximation error (SAE). Their studies show us that Sparse Codes (SCs) are highly informative in ECG classification and more easily adaptable to changes in ECG patterns. It appears that SC-based arrhythmia monitoring has the potential for significant improvements.

In this study, we begin with the approach in \cite{carrera2016ecg, Elad}, and \cite{carrera2019online} to learn personalized dictionaries representative of normal beats. We investigate novel ways to improve anomaly detection performance by instrumenting advanced machine learning tools without using any abnormal beats and only a small amount of normal beats for a specific user. The novel and significant contributions of this study can be introduced as follows. (i) First, \textit{we introduce the left null space matrix of the normal dictionary. This provides a faster way to obtain approximated representation error similar to the one proposed in \cite{carrera2016ecg, carrera2019online}}. The proposed null space analysis-based classifier performs very similarly to the SA analysis in \cite{carrera2016ecg, carrera2019online} with a significantly reduced computational cost. Additionally, we analyze how the classifier based on the regularized least square solution can be used as an alternative to the null space-based classifier, with a similar classification accuracy and computational complexity.
\textit{(ii) A domain transformation technique is presented in this study for projecting the heartbeats of a patient to the space of a new user's heartbeats without knowledge of the new user's abnormal data.} Different from \cite{kiranyaz2017personalized}, the transformation is linearly approximated in this case and the linear mapping is estimated using sparse representation where only the normal heartbeat dictionaries of the target (i.e. new users with no abnormal heartbeats, but a small normal heartbeat collection) and source (i.e. registered patients with both normal and abnormal heartbeats) users are used. The anomaly detection performance is significantly improved compared to the SOTA methodology proposed in \cite{kiranyaz2017personalized} when the same classifier is trained and used.

\textit{(iii) Moreover, by combining innovations (i) and (ii), we developed a decision-making strategy that enhanced SOTA performance even further.} \textit{(iv) Finally, a novel and energy-efficient ECG monitoring system is proposed to detect arrhythmia at the moment it occurs.} In this system, the ECG is first projected onto the left null space of the normal dictionary using matrix-vector multiplication, which requires insignificant power. Then, ECG beats are pre-classified (in real-time and possibly on the sensor) by using the energy of these projected signals, and only the suspected ones are further classified using the proposed methods in (ii) or (iii). Thus, using this lightweight monitoring scheme, the energy of the sensors can be preserved considerably with minimal sacrifice on the detection performance.

The rest of the paper is organized as follows. In Section \ref{sec:notations}, the notations are introduced. A brief literature survey is presented in Section \ref{sec:literature}. The proposed null space projection-based classification and sparse representation-based domain adaptation are explained in Section \ref{sec:proposed}. In Section \ref{sec:results}, experimental results and comparative evaluations are presented. Finally, Section \ref{sec:discussion} and Section \ref{sec:conclusion} discuss the performance and limitations of the proposed system with possible future extensions and highlight important conclusions.

\subsection{Notations}
\label{sec:notations}
Throughout the article, the $\ell_p$-norm of a vector $\bm{x} \in \mathbb{R}^N$ is defined as $\left\| \bm{x} \right\|_{p} = (  \sum_{i=1}^N \left\vert x_i \right\vert^p )^{1/p}$ for $p \geq 1$. In addition, the number of non-zero coefficients of $\bm{x}$ is calculated with the $\ell_0$-norm, i.e., $\left\| \bm{x}  \right\|_{0} = \lim_{p \to 0} \sum_{i=1}^N \left\vert x_i \right\vert^p = \# \{ l: x_l \neq 0 \}$. A strictly $k$-sparse signal contains at most $k$ number of non-zero coefficients, i.e., $\left\|\bm{x} \right\|_{0} \leq k$. As a convenience to readers, we provide a list of frequently used variables, their definitions, and their abbreviations in Table \ref{tab:definitions}.

\begin{table}
    \captionsetup{font=footnotesize}
    \centering
    {
    \setlength\tabcolsep{3pt}
    \renewcommand{\arraystretch}{1.2}
    \resizebox{\columnwidth}{!}{\begin{tabular}{l|l|l}
    \hline
    \rowcolor{gray!30} \textbf{Variable} & \textbf{Synonyms} & \textbf{Properties} \\
    \hline
    $\bm{s_i^p} \in \mathbb{R}^N$ & \makecell[l]{ $i^{th}$ ECG beat of \\ user $p$} & Sparse in $\bm{D}$ \\
    \hline
    $\bm{D} \in \mathbb{R}^{N \times n}$ & Sparsifying dictionary & $\bm{s_i^p} = \bm{D^px_i^p} + \bm{e}$ \\
    \hline
    $\bm{e} \in \mathbb{R}^N$ & Representation error & \\
    \hline
    $\bm{x_i^p} \in \mathbb{R}^n$ & Sparse coefficients of $\bm{s_i^p}$ & $\left\| \bm{x} \right\|_0 \leq k$ if $k$-sparse \\
    \hline
    $\bm{\widetilde{e}_{\ell_1}} = \bm{D^p\widehat{x}_i^p} - \bm{s_i^p}$ &\makecell[l]{Sparse Approximation \\ Error (SAE)} & \makecell[l]{$\bm{\widehat{x}_i^p}$ is the solution of \\ any SR algorithm} \\
    \hline
    $\bm{F^p} \in \mathbb{R}^{N - n \times N}$ & Left annihilator matrix & $\bm{F^pD^p} = 0$ \\
    \hline
    $\bm{\widetilde{e}} = \bm{F^ps_i^p}$ & \makecell[l]{Null Space Projection \\ Error (NPE)} & \\
    \hline
    $\bm{\widetilde{e}_{\ell_2}} = \bm{D^p \widehat{x}_i^p} - \bm{s_i^p}$ & \makecell[l]{Least Squares Approximation \\ Error (LAE)} & \makecell[l]{$\bm{\widehat{x}_i^p}$ is the Least \\ Squares estimate} \\
    \hline
    $\bm{Q_{l \rightarrow p}} \in \mathbb{R}^{N \times N}$ & \makecell[l]{Morphology Transformation \\ Matrix (MTM)} & \makecell[l]{ECG beat linear domain \\ adaptation matrix from \\ user $l$ to user $p$} \\
    \hline
    \end{tabular}}
    }
    \caption{Frequently used variables throughout the article.}
    \label{tab:definitions}
\end{table}

\section{Related Works}
\label{sec:literature}
\subsection{Sparse Representation Based Classification}
\label{sec:SAE}
Let $\bm{s_{i}^p} \in \mathbb{R}^N$ be the $i^{th}$ ECG heartbeat of the $p^{th}$ user which can be represented \cite{carrera2016ecg} by a linear combination of $n$ wave-forms, i.e.,
\begin{equation}
\bm{s^p_i}=   \bm{{D^p}} \bm{x^p_i}, \label{eq:gen_rep}
\end{equation}
where $\bm{D^p} \in \mathbb{R}^{N \times n} $ is the dictionary of these wave-forms (atoms) and $\bm{x^p_i} \in \mathbb{R}^{n}$ is the corresponding coefficient vector. In the case of being overcomplete ($n > N$) the representation will be enriched; however, the representation formed by \eqref{eq:gen_rep} will not be unique. It is possible to make it unique by making it represent the signal in $\bm{D^p}$ with the smallest number of significant coefficients (i.e. sparseness), i.e., 
\begin{equation}
\min_{\bm{x^p_i}} ~ \left\| \bm{x^p_i} \right\|_{\ell_0^n}~ \text{subject to}~ \bm{D^p} \bm{x^p_i}=\bm{s^p_i} ,\label{eq:sparse_rep}
\end{equation}
which is known as the sparse representation of the signal $\bm{s^p_i}$. A signal is said to be strictly $k$-sparse if the number of non-zero coefficients representing it in the dictionary is less than a constant $k$, i.e., $\left\| \bm{x} \right\|_0 \leq k$. The most common solutions for solving the sparse optimization problem are convex relaxation, like Basis Pursuit and Basis Pursuit Denoising \cite{BP}, or greedy algorithms like orthogonal matching pursuit (OMP) \cite{OMP1}.

The sparse representation in \eqref{eq:sparse_rep} is still valid in under-complete dictionaries i.e, $N>n$. For instance, the authors of \cite{carrera2016ecg} use $n$ number of atoms with $n<N$, e.g., $n = 20$ where $N = 256$, and get the highest performance in representing the normal ECG heartbeats. The authors of \cite{carrera2016ecg} use the energy of \textbf{Sparse Approximation Error (SAE)} to detect anomalies, which can be calculated as
\begin{equation}
    r_i = \left\| \bm{D^p} {\bm{\hat{x}^p_i}}-\bm{s^p_i} \right\|_2^2, \label{eq:residual}
\end{equation}
where $\bm{D^p}$ is the user-specific dictionary learned by using the KSVD method \cite{KSVD} from the normal heartbeats. Normal beats are assumed to be collected for a limited duration from a new user without obtaining or labeling the abnormal beats. Having the pre-trained dictionary, when a new ECG beat $\bm{s^p_i}$ comes into the picture, its sparse coefficients are obtained with the OMP algorithm. Then, the estimation $\bm{\hat{x}^p_i}$ is back-projected to the dictionary and the residual or SAE is calculated as in Eq. \eqref{eq:residual}. Then, they use a simple thresholding method on $r_i$ to detect anomaly i.e., detect anomaly if $r_i>\gamma$, where $\gamma$ is the defined threshold.

\subsection{Abnormal Beat Synthesis for Early Arrhythmia Detection}
\label{sec:ABS}
Using both normal and abnormal ECG beats collected from the patient, an anomaly ECG beat detection system can easily be trained. The problem, however, is that in real life, if we want to develop an early warning system, we may never have the real anomaly beats for a healthy user who has just registered to the system. To resolve this problem, the authors of \cite{kiranyaz2017personalized} introduced a personalized abnormal beat synthesis (ABS) method. In this system, when a new user enters the system, the normal ECG beats of this new user are collected for a short period of time. Since the assumption is that there are no anomalous heartbeats in these signals, synthetic anomaly heartbeats are generated from them.

To develop their system, the authors first modeled the effect of common causes of heart disease on the measurement of ECG beat signals. In this model, a certain group of heart diseases morphologically distort normal heartbeat signals to convert them into a certain type of abnormal heartbeat signals. The authors modeled this transformation from a normal heartbeat to an abnormal heartbeat as a linear time-invariant (LTI) system. Mathematically speaking, the assumption is that any observed abnormal signal $\bm{s_A^l}$ of user $l$ can be approximated as the linearly degraded version of a latent unknown normal signal $\bm{s_N^l}$, i.e.,
\begin{equation}
    \bm{s_N^l} \circledast \bm{h^l} = \bm{s_A^l},
\end{equation}
where $\bm{h^l} \in \mathbb{R}^M$ is the $M$-length filter coefficient vector of the LTI system and $\circledast$ is the convolution operation. As part of the process of obtaining a filter-bank, the authors first select an average normal beat for a user $l$, by taking the average of all normal beats of the user and finding the normal beat closest to the average. Then, using this average normal beat $\bm{\bar{s}_N^l}$, for each abnormal observation $\bm{s_A^l}$, a filter $\bm{h^l}$ is estimated from Equation $\bm{\bar{s}_N^l} \circledast \bm{h^l} = \bm{s_A^l}$ via regularized least-squares filtering. The final \textbf{ABS} filter library is the remaining set of filters after similar ones have been pruned.

When a new user $p$ registers to the system, the average normal beat $\bm{\bar{s}_N^p}$ of this new user is calculated from the collected beats. Then, a kernel is picked up from the aforementioned filter bank of some user $l$. Finally, a synthetic abnormal beat $\bm{s_A^p}$ is generated by applying this filter to $\bm{\bar{s}_N^p}$ as $\bm{s_A^p} = \bm{\bar{s}_N^p} \circledast \bm{h^l} $. This way, the personalized training data for user $p$ is generated as a collection of $p$'s own normal beats and generated abnormal beats using the ABS filter library. 
Such artificial generation of the  personalized training data enables the training of an ordinary classifier, e.g., a 1-D CNN. Figure \ref{fig:linear_degrading_system} illustrates the use of linear degradation estimates of an existing user on a new user's normal ECG signals.


\input{figures_tikz/linear_degrading_system}

\subsection{Generative Adversarial Network Based Synthesizers }
Typically, certain arrhythmia types occasionally occur in an ECG recording acquired even from arrhythmia patients. Most of the GAN-based ECG generator algorithms \cite{golany2019pgans,cab, ODE-GAN, simgans, wang2019ecg, shaker2020generalization}
were developed to provide data augmentation so as to  overcome the data imbalance problem during training rather than to provide a \textit{personalized zero-shot} (personalized zero-shot refers to being entirely blind to the person's anomaly signal during training) solution. In the sequel, a brief overview of the developments in GAN-based ECG data generation technologies is presented. 


In \cite{delaney2019synthesis}, Delaney \textit{et al.} trained various GANs to generate synthetic signals and used dynamic time warping (DTW) and maximum mean discrepancy (MMD) to evaluate the quality of the generated signals. Similar to \cite{delaney2019synthesis}, \cite{wulan2020generating} used GANs for data generation and evaluated their performances by using an SVM classifier. Zhu \textit{et al.} \cite{zhu2019electrocardiogram} used GANs with bidirectional LSTM-CNNs for data augmentation and used percent root mean square distance (PRD), root mean square error (RMSE), and Fréchet distance (FD) for synthesizer evaluation. In these approaches, the aim was once again to generate synthetic ECG data, hence they neither train a classifier nor test the classification performance.

In \cite{zhou2021fully}, Zhou \textit{et al.} proposed the most akin method to ours, which can therefore be categorized as personalized zero-shot learning. The authors trained a global generative network, where the generator synthesizes a 2-D coupling matrix belonging to one of five beat classes, and the discriminator works as an auxiliary classifier to determine both the beat class and its genuinity. After the initial training, the discriminator is fine-tuned into a patient-dependent classifier by further training on normal beats and by generating abnormal beats from a specific patient.

Among the aforementioned GAN-based ECG generator algorithms that try to overcome the data imbalance problem, we select one of the SOTA methods \cite{shaker2020generalization} as a competing algorithm even though a direct comparison is not fair since in our zero-shot setup we do not have access to any anomaly beat of the user while they do. Similar to \cite{zhou2021fully}, Shaker et al. \cite{shaker2020generalization} trained a global GAN synthesizer for data augmentation; however, they trained a separate classifier for arrhythmia detection. Even though this method is neither zero-shot nor personalized (which brings advantages compared to our zero-shot solution in any comparison), we still compare our performance as it can be intuitively seen as an upper band for our task of data generation for the case where there is access to at least a few anomalies for the patient. Despite such disadvantages, the proposed solution in this study can still achieve comparable results as discussed in Section VI.

\section{Proposed Solution(s)}
\label{sec:proposed}
\subsection{Null Space Projection Based Classifier (Proposed Solution I)} \label{sec:NPE}
\begin{figure}[!htbp]
    \captionsetup{font=footnotesize}
    \centering
    \includegraphics[draft=false,width=0.47\textwidth]{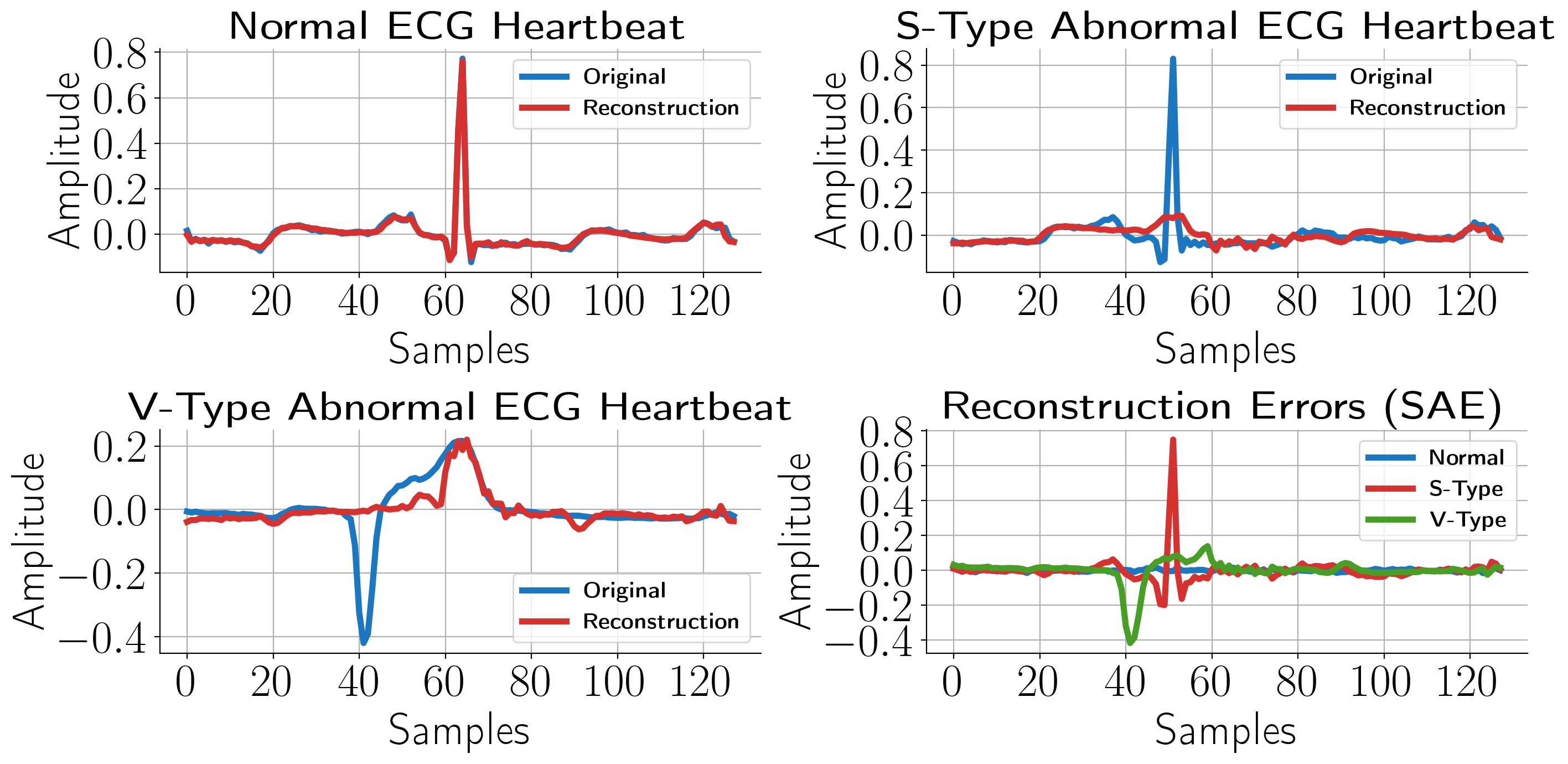}
    \caption{Example of normal, S-type, and V-type ECG signals from patient 100, and their sparse approximation errors. From \ref{fig:beat_types_and_errors}d it is clear that both types of abnormal beats have noticeably large SAE compared to a normal beat.}
    \label{fig:beat_types_and_errors}
\end{figure}

As in \cite{carrera2016ecg}, we also assume that a normal ECG beat can be represented in a pre-defined dictionary representing normal beat space with a relatively small approximation error, i.e., 
\begin{equation}
\bm{s^p_i} = \bm{{D^p}} \bm{x^p_i} + \bm{e}, \label{eq:gen_rep_w_error}
\end{equation}
where $\bm{e}$ can be called as representation error, or residual (later will be called as SAE when $\bm{x^p_i}$ is sparse). For the ideal case where $\bm{e} = 0$ and $\bm{{D^p}}$ satisfies some properties, the uniqueness of the solution of \eqref{eq:gen_rep_w_error} can be guaranteed \cite{RIP} with SR given in Eq. \eqref{eq:sparse_rep} with an assumption that $\bm{x^p_i}$ is strictly sparse. In a real-world ECG classification task, one can easily assume that sparse approximation errors, $\bm{e}$, always occur, and the representation coefficients, $\bm{x}$ are not exactly but approximately sparse. Moreover, as the $\ell_0$-norm is a non-convex function, the optimization problem in Eq. \eqref{eq:sparse_rep} is NP-hard. By replacing the $\ell_0$-norm with the closed convex norm, $\ell_1$-norm, this $\ell_0$-minimization problem can be relaxed:   
\begin{equation}
    \min_{\bm{x^p_i}} \left\| \bm{s^p_i}-  \bm{{D^p}} \bm{x^p_i} \right\|_2^2 + \lambda \left\| \bm{x^p_i} \right\|_1, \label{eq:l1-minimization}
\end{equation}
which is also known as the Lasso formulation \cite{lasso}, and known to provide a stable solution in noisy and approximate sparse cases, and a stable solution in noise-free cases \cite{lasso-stable} given the dictionary satisfies some properties. There exist numerous techniques in order to handle the optimization problem defined in Eq.~\eqref{eq:l1-minimization}. We implemented the proposed algorithm based on the alternating direction method of multipliers (ADMM) \cite{ADMM}. As can be seen in Figure \ref{fig:beat_types_and_errors}, a normal ECG beat $\bm{s^p_i}$ can be well represented in $\bm{{D^p}} $, i.e., $\bm{s^p_i} \approx \bm{{D^p}} \bm{x^p_i}$, while abnormal ones having larger representation error when represented in $\bm{{D^p}} $.

\underline{Dictionary Learning:} As stated in Section \ref{sec:SAE}, by using the nice properties of sparse representation, in the methods \cite{carrera2016ecg, carrera2019online}, the user-specific dictionary $\bm{{D^p}}$ is learned only from the normal ECG beat collection of user $p$. Different from their work, where K-SVD is used to learn dictionaries for a pre-determined fixed number of non-zero coefficients, $k$, we use the following Lasso formulation, in which the estimated sparse signal coefficients have a more flexible structural behavior:
\begin{equation}
    \min_{\bm{X^p}, ~\bm{{D^p}}} \left\| \bm{S^p}-  \bm{{D^p}} \bm{X^p} \right\|_2^2 + \lambda \left\| \bm{X^p} \right\|_1, \label{eq:dictionary_learning}
\end{equation}
where $\bm{S^p} \in \mathbb{R}^{N \times T}$ is the collection of T number of normal ECG beats of user $p$ and $\bm{X^p} \in \mathbb{R}^{n \times T}$ is the corresponding sparse coefficient matrix. In order to solve the optimization problem defined in Eq. \eqref{eq:dictionary_learning}, we implement a variant of the method of optimal directions (MOD) \cite{MOD}. All ECG beat samples considered in the study have been used only after normalized to the unit norm, i.e.,  $ \left\| \bm{s_i} \right\|_2 = 1$.

\underline{Energy-Efficient Classification:}
Having the dictionary $\bm{{D^p}}$ of the healthy ECG beat space, when a new set of test signals $\bm{S^p_i}$ are acquired, one can solve the sparse recovery problem in Eq. \eqref{eq:l1-minimization}. Then, by using the SAE outlined in Eq. \eqref{eq:residual}, the class can be determined using a predefined threshold \cite{carrera2016ecg}. Besides the severe threshold sensitivity, such sparse recovery algorithms still work in an iterative manner, which increases the computational cost.

On the other hand, leveraging only a few atoms (e.g., $n=20$) compared to the beat signal size (e.g., $N=128$) is enough to represent normal signal (beat) space. In such a scenario (i.e., $N>n$), the left null space of $\bm{{D^p}}$ exists and, in the sequel we will show that by projecting the test signal on it, the computational complexity of the classification task can be reduced significantly. Let $\bm{F^p} \in \mathbb{R}^{N-n \times N}$  be the left annihilator matrix of $\bm{{D^p}}$, i.e., $\bm{F^p} \bm{{D^p}}=0$. We find such a matrix by orthonormalizing the left-null space basis of $\bm{{D^p}}$.

Given the test signal, $\bm{s^p_i}$, the pre-constructed and user-specific normal signal component annihilator matrix, $\bm{F^p}$, is applied from the left, i.e., $\bm{F^p}  \bm{s^p_i}$. By using Eq. \eqref{eq:gen_rep_w_error}, the resulting vector $\widetilde{\bm{e}}$, which we call \textbf{Null Space Projection Error (NPE)}, can be cast as
\begin{equation}
\widetilde{\bm{e}} = \bm{F^p} \bm{s^p_i}=   \bm{F^p} \left(  \bm{{D^p}} \bm{x^p_i} + \bm{e} \right) = \bm{F^p} \bm{e}. \label{eq:NPE}
\end{equation}

By doing so, the normal ECG beat components can be removed through simple matrix-vector multiplication. It is straightforward to see that the energy of NPE would be small enough for a healthy ECG signal, but would be much greater in the case of an abnormal signal. On the other hand, we will demonstrate in Section \ref{sec:results} that the energy levels of both SAE and NPE are usually very close when applied to real ECG beats. But compared to the traditional approach where the sparse coefficients are estimated first by $\ell_1$-minimization, and then the SAE is computed, the computational complexity of the NPE is significantly low.  

\begin{figure}[!htbp]
    \captionsetup{font=footnotesize}
    \centering
    \includegraphics[draft=false,width=0.47\textwidth]{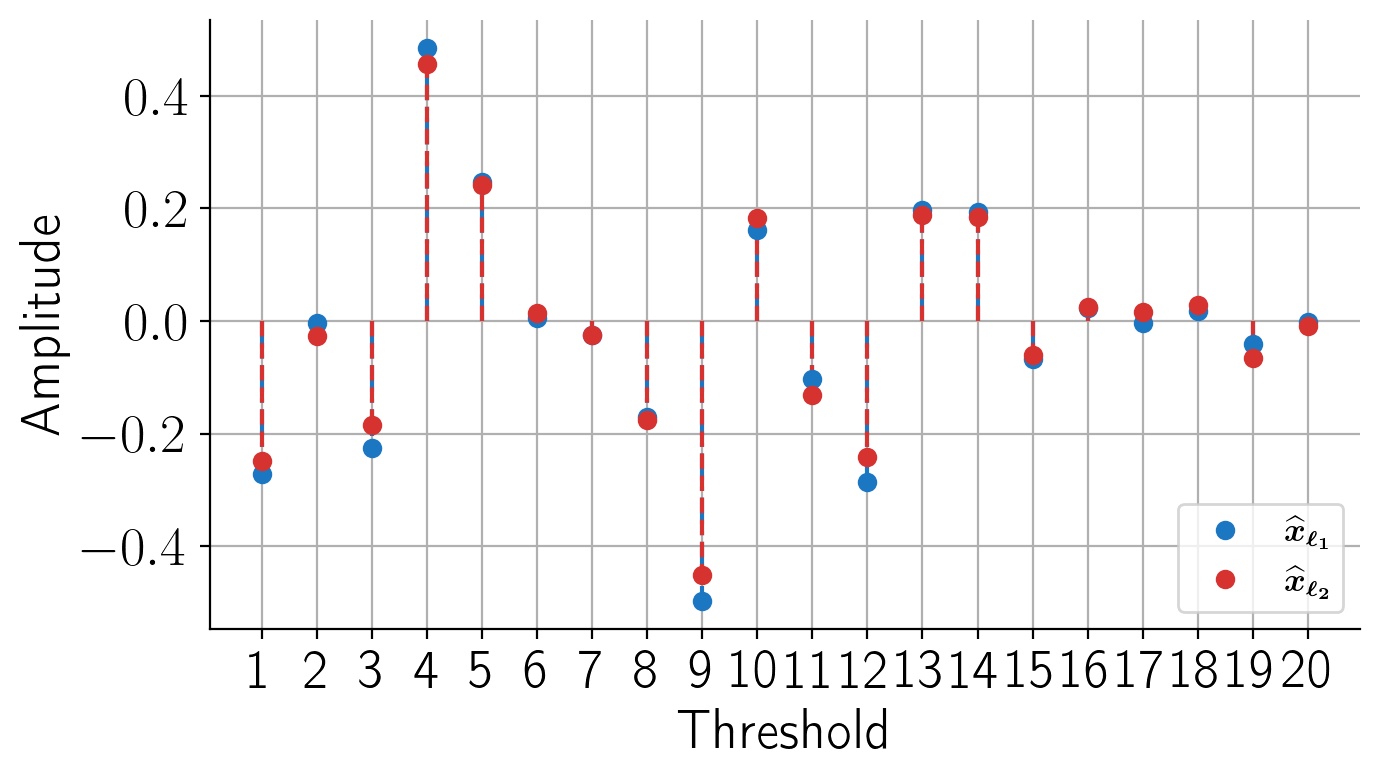}
    \caption{Sparse codes calculated through $\ell_1$ and $\ell_2$ minimization for a normal ECG beat and a dictionary with $n=20$ atoms.}
    \label{fig:sparse_codes_l1_l2}
\end{figure}

\subsection{Classification by Least-Squares Approximation as An Alternative Solution} \label{sec:LAE}

SAE-based classification is a special case of the well-known sparse representation-based classification (SRC) \cite{SRC1}. The original SRC targets a multi-class classification problem. After SRC gained popularity, the authors of \cite{collaborative} proved that in many multi-class classification problems, similar classification performance to SRC can be achieved by using $\ell_2$ regularization instead of $\ell_0$ or $\ell_1$. The approach is called Collaborative Representation based Classification (CRC) and can significantly reduce the computational complexity compared to SRC.

By adopting a similar strategy, we can swap the $l_1$ term in Eq. \eqref{eq:l1-minimization} with $\lambda \left\| \bm{x_i^p} \right\|_2^2$ to get the ridge formulation of the problem and solve it using the method of regularized least squares: 
\begin{equation} \label{eq:l2-solution}
    \bm{\widehat{x}_i^p} = \underbrace{\left({\bm{D^p}}^T \bm{D^p} + \lambda \bm{I}_{N \times N} \right)^{-1}{\bm{D^p}}^T}_{\bm{L^p}} \bm{s_i^p}
\end{equation}
where $\bm{L^p} = \left({\bm{D^p}}^T \bm{D^p} + \lambda \bm{I}_{N \times N} \right)^{-1}{\bm{D^p}}^T $ and $\lambda$ is a small positive scalar. Having the estimation $\bm{\widehat{x}_i^p} $, \textbf{the Least-Squares Approximation Error (LAE)} can be calculated easily as $\bm{\widetilde{e}_{\ell_2}} = \bm{D^p} {\bm{\hat{x}^p_i}}-\bm{s^p_i} $. We will show that LAE and NPE require significantly less computational cost when compared to SAE, and provide almost identical classification results with the simple thresholding-based classifier. In Figure \ref{fig:sparse_codes_l1_l2}, an example pair of estimated representation coefficients with $\ell_1$ and $\ell_2$ regularizes are given respectively. Although $\ell_1$-minimization can provide a slightly sparser (more ideal) estimation of the representation coefficient vector, the energy of the representation errors (SAE, LAE) are almost identical for this specific task.

\underline{Complexity Analysis:} Given the pre-constructed annihilator matrix, $\bm{F^p} \in \mathbb{R}^{(N - n) \times N}$, and a single ECG beat, $\bm{s_i^p} \in \mathbb{R}^{N}$, the NPE can be calculated through matrix-vector multiplication. Thus, the exact number of required floating-point operations (FLOPs) for calculating the NPE is $T_{\text{NPE}}(N, n) = 2 \cdot N \cdot \left(N - n\right)$ FLOPs.

When calculating the LAE, as $\bm{D^p}$ is calculated beforehand, we can pre-construct $\bm{L^p}$ defined in Eq \ref{eq:l2-solution}. Then, the Least-Squares Approximation Error (LAE) is expressed as,
\begin{equation} \label{eq:lae}
    \bm{\widetilde{e}_{\ell_2}} = \bm{s_i^p - D^p L^p s_i^p},
\end{equation}
which involves a matrix-vector multiplication of $\bm{L^p} \in \mathbb{R}^{n \times N}$ and $\bm{s_i^p} \in \mathbb{R}^{N}$, followed by another matrix-vector multiplication of $\bm{D^p} \in \mathbb{R}^{N \times n}$ and the intermediate result, and finally a subtraction with the original signal. Thus, the computational complexity of calculating the LAE is $T_{\text{LAE}}(N, n) = \left(4 \cdot n + 1\right) \cdot N$ FLOPs. It is worth noting that one could also pre-construct $\bm{I_{N\times N}}- \bm{D^p L^p} $ in Eq. \eqref{eq:lae} and calculate LAE with a single matrix-vector multiplication, i.e., $\bm{\widetilde{e}_{\ell_2}} = \left( \bm{I_{N\times N}} - \bm{D^p L^p} \right) \bm{s_i^p} $. In that case, the computational complexity would be $T_{\text{LAE}}(N) = 2 \cdot N^2$ FLOPs. The more suitable approach can be chosen depending on $N$ and $n$, and the computational platform. Although the first approach may require fewer FLOPs, the second approach may be more efficient when it comes to computation time since all the inner products can be computed in a parallel manner.

As multiple factors play a role in the number of FLOPs needed to calculate the SAE, such as the sparse recovery algorithm and the number of iterations, calculating the exact number of FLOPs is more challenging. In \cite{carrera2016ecg}, OMP was used to find sparse codes, following which these sparse codes were used to estimate the sparse approximation error, i.e., $\bm{\widetilde{e}_{\ell_1}} = \bm{D^p} {\bm{\hat{x}^p_i}}-\bm{s^p_i}$. In their algorithm, the authors of \cite{carrera2016ecg} also fixed the number of non-zero coefficients of $\bm{\hat{x}^p_i}$ to a predetermined number (e.g., $k=5$). In this way, the number of iterations in the OMP algorithm is fixed, therefore the number of FLOPs required for OMP can be approximated as $2 \cdot N \cdot k \cdot \left(k + 1.5  \right) + 2 \cdot k \cdot n \cdot \left(N + 1 \right)$. Having the sparse codes, SAE can be calculated in $(2n+1) N$ FLOPs. In Tables \ref{tab:FLOPs} and \ref{tab:runtimes}, the number of required FLOPs for different approximation errors and their execution times are listed.

When a simple thresholding algorithm is applied to the energy of these error vectors, namely SAE, NPE, and LAE, they result in almost the same AUC performance.
Among them, the NPE achieves the lowest computational complexity especially when parallel processing units are available. 

\begin{table}
    \captionsetup{font=footnotesize}
    \centering
    {
    \setlength\tabcolsep{3pt}
    \renewcommand{\arraystretch}{1.2}
    \begin{tabular}{l|l|l}
    \hline
    \rowcolor{gray!30} \textbf{Error} & \textbf{AUC} & \textbf{Complexity (FLOPs)} \\
    \hline
    SAE & 0.97019 & \makecell[tl]{$\approx 2 \cdot N \cdot k \cdot \left(k + 1.5 \right) +$ \\ \quad $2 \cdot k \cdot n \cdot \left(N + 1 \right) + (2 \cdot n+1) \cdot N$} \\
    NPE & 0.96993 & $2 \cdot N \cdot (N - n)$ \\
    LAE & 0.97002 & $\left(i\right)$ $2 \cdot N^2$ or $\left(ii\right)$ $\left(4 \cdot n + 1 \right) \cdot N$ \\
    \hline
    \end{tabular}
    }
    \caption{Number of FLOPs needed for calculating different approximation errors.}
    \label{tab:FLOPs}
\end{table}

\begin{table}
    \captionsetup{font=footnotesize}
    \centering
    {
    \setlength\tabcolsep{3pt}
    \renewcommand{\arraystretch}{1.2}
    \begin{tabular}{c|c|c}
    \hline
    \rowcolor{gray!30} \textbf{Error} & \textbf{CPU Runtime ($\mu$s)} & \textbf{GPU Runtime ($\mu$s)} \\
    \hline
    SAE & 4.7933 & 11.6808 \\
    NPE & 0.2047 & 0.1032 \\
    LAE $\left(1\right)$ & 0.2498 & 0.1062 \\
    LAE $\left(2\right)$ & 0.2133 & 0.1196 \\
    CNN & 21.0028 & 9.0496 \\
    \hline
    \end{tabular}
    }
    \caption{The processing time of a single ECG beat for the discussed methods on an i7-10870H CPU and an RTX 3080 Laptop GPU. The measurements are averaged over 10 runs and 1000 beats.}
    \label{tab:runtimes}
\end{table}

\subsection{Sparse Representation Based Domain Adaptation}
\label{sec:SR-DA}
As it has been mentioned several times in past studies, ECG morphology differs between individuals depending on their cardiovascular systems \cite{ECGDataset1}. This is why recent research on biometrics of ECG signals has gained popularity \cite{ECGBiometrics1, ibtehaz2021edith}. The authors of \cite{kiranyaz2017personalized} studied how a normal heartbeat signal can be transformed into an abnormal one, and found that this transformation can be approximated with a linear transformation. Using this work as an inspiration, we investigate for the first time in the literature the relationship between the ECG morphology of user $l$ and $p$. In the sequel, we will demonstrate that such a relationship can also be modeled accurately enough with a linear transformation.

Regardless of whether it is an anomaly or not, when an ECG heartbeat signal $\bm{s_i^l}$ of user $l$ is linearly transformed, i.e, $\bm{\widehat{s}_i^l} = \bm{Q_{l \rightarrow p}} \bm{s}_i^l$, we want the transformed beat $\bm{\widehat{s}_i^l}$ to have the same morphology as the real ECG beats of user $p$. We would like to emphasize that this \textbf{morphology transformation matrix (MTM)}, $\bm{Q_{l \rightarrow p}}  \in \mathbb{R}^{N \times N}$, pertains specifically to the transition from the ECG heartbeat domain of user $l$ to the one of user $p$. Such a transformation system is illustrated in Figure \ref{fig:linear_transformation_system}.

\begin{figure}[!htbp]
    \captionsetup{font=footnotesize}
    \centering
    \includegraphics[draft=false,width=0.47\textwidth]{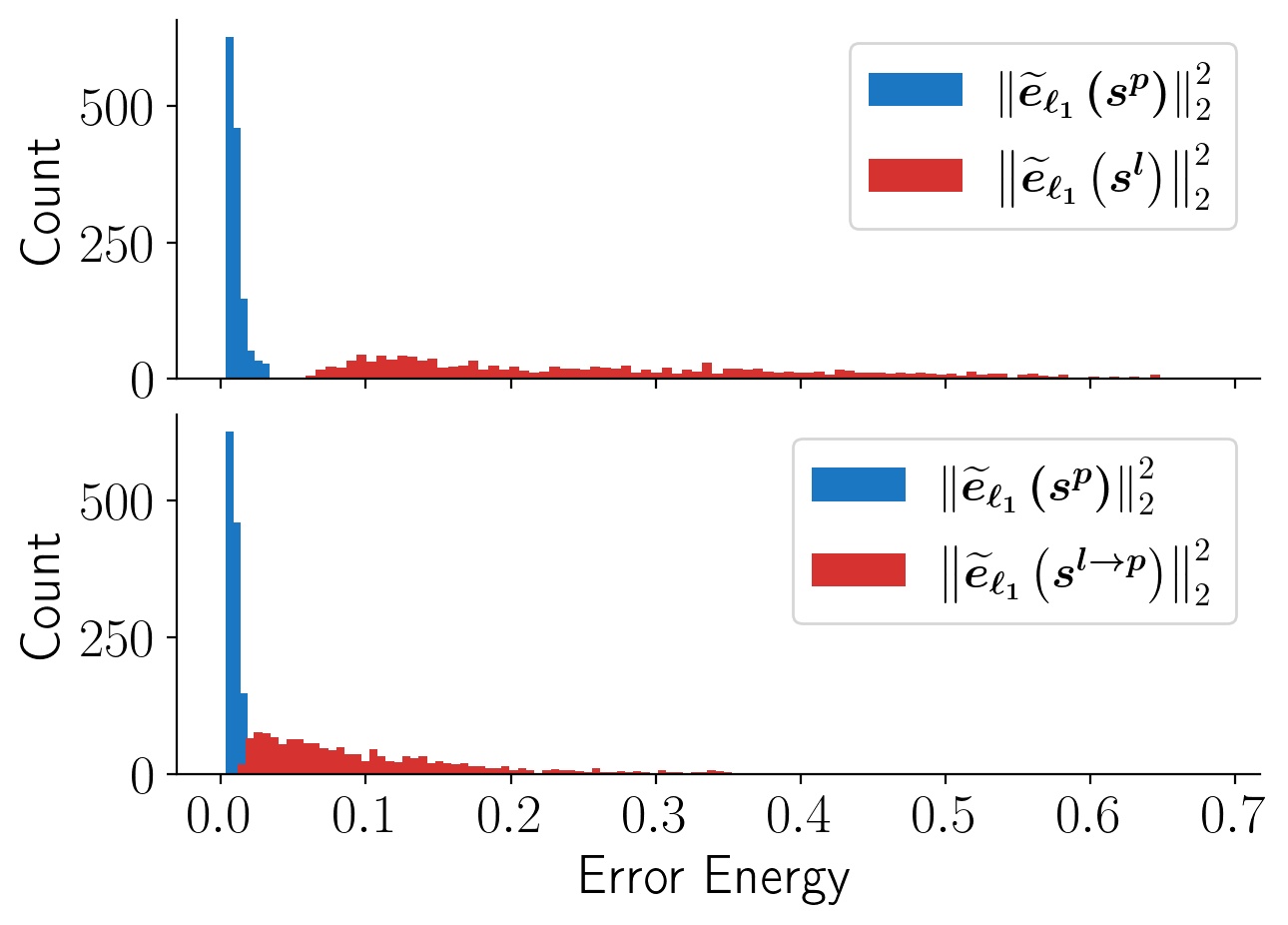}
    \caption{The histogram of sparse approximation errors of normal beats of users $\bm{p}$ and $\bm{l}$ on user $\bm{p}$'s dictionary, $\bm{D^p}$. $\bm{\widetilde{e}_{\ell_1}}$ for $\bm{l}$ is high even for normal beats on $\bm{D^p}$ (upper), whereas after domain adaptation the error energy more closely resembles the ones of $\bm{p}$ (lower).}
    \label{fig:SAE_of_p_and_l_on_D_after_DA}
\end{figure}

Such transformation matrices are challenging to learn since there are only a limited number of healthy beats with no abnormal data available for user $p$. This may lead any learning algorithm to overfit to this small number of samples. If we consider the concept of the sparse representation-based dictionary learning as shown in Eq. \eqref{eq:dictionary_learning}, we should assume that when a set of real heartbeats, $\bm{S^p}$, belonging to user $p$ arrive, this set will be represented in a profound (as a result of dictionary learning defined in Eq. \eqref{eq:dictionary_learning}) dictionary $\bm{D^p}$ with relatively small errors in representation. Nonetheless, as shown in Figure \ref{fig:SAE_of_p_and_l_on_D_after_DA}a, if we acquire a new set of healthy signals, $\bm{S^l}$, belonging to a different user such as user $l$, we can assume that this set will not be well represented in the dictionary of user $p$ despite the absence of abnormal beats.

\input{figures_tikz/linear_transformation_system}

Based on the discussion above, we formulate our MTM learning problem as finding a transformation matrix, $\bm{Q_{l \rightarrow p}}  \in \mathbb{R}^{N \times N}$, which results in the $\bm{S^l}$ being sparsely represented in $\bm{D^p}$ after transformation: 
\begin{multline}
    \bm{\hat{Q}_{l \rightarrow p}}, \bm{\hat{X}^l} = \underset{\bm{Q_{l \rightarrow p}},\,\bm{X^l}}{\arg\min} \|\bm{Q_{l \rightarrow p}S^l} - \bm{D^pX^l}\|_2^2 \\ + \lambda \|\bm{X^l}\|_1 + \gamma\|\bm{S^l} - \bm{Q_{l \rightarrow p}S^l}\|_2^2 \label{eq:domain_adaptation}
\end{multline}
where $\lambda$ and $\gamma$ are positive hyper-parameters and the last term, $\gamma\|\bm{S^l} - \bm{Q_{l \rightarrow p}S^l}\|_2^2$, is a trade-off term. For large values of $\gamma$, it ensures that $\bm{\widehat{S}^l}$ does not deviate much from $\bm{S^l}$. For small values of $\gamma$, $\bm{\widehat{S}^l}$ fits better onto dictionary $\bm{D^p}$. Without the last term, or for $\gamma = 0$, Eq. \eqref{eq:domain_adaptation} has a trivial minimum at $\bm{X^l} = 0$ and $\bm{Q_{l \rightarrow p} = 0}$. In other words, Eq \eqref{eq:domain_adaptation} solves for a mapping that compromises between projecting the signal onto subspace $\bm{D^p}$ and remaining close to the original signal.

In order to solve the problem defined in Eq. \eqref{eq:domain_adaptation}, we follow an iterative optimization strategy: Given $\bm{D^p}$ and $\bm{S^l}$ at each iteration, first, the current solution for $\bm{Q_{l \rightarrow p}}$ is fixed and the sparse codes $\bm{X}$ are estimated; then, the solution for $\bm{X}$ is fixed and $\bm{Q_{l \rightarrow p}}$ is updated. We would like to remind that the ECG beats and dictionary atoms are always normalized to have unit energy, i.e., $\left\| \bm{s^l_i} \right\| = \left\| \bm{d_i} \right\| = \left\|\bm{\widehat{s}^l_i}\right\|=1$. 

At each iteration, given the current solution $\bm{Q_{l \rightarrow p}}$, each signal $\bm{s_i^l}$ in the set $\bm{S^l}$ is linearly projected onto $\bm{Q_{l \rightarrow p}}$, i.e., $\bm{\widehat{s}^l_i} =\bm{Q_{l \rightarrow p}} \bm{s^l_i}$, and then normalized. In order to estimate $\bm{X}$, Eq. \eqref{eq:domain_adaptation} can be reduced to
\begin{equation}
    \bm{\hat{X}^l} \gets \underset{\bm{X^l}}{\arg\min} \left\| \bm{\widehat{S}^l} - \bm{D^p} \bm{X^l} \right\|_2^2 + \lambda \left\| \bm{X^l} \right\|_1.
\end{equation}
Then, for a fixed $\bm{X}$, the optimization problem in Eq. \eqref{eq:domain_adaptation} is recast as
\begin{multline}
    \bm{\hat{Q}_{l \rightarrow p}} = \underset{\bm{Q_{l \rightarrow p}}}{\arg\min} \|\bm{Q_{l \rightarrow p}S^l} - \bm{D^pX^l}\|_2^2 \\ + \gamma\|\bm{S^l} - \bm{Q_{l \rightarrow p}S^l}\|_2^2. \label{eq:quadratic}
\end{multline}
The function to be minimized in Eq. \eqref{eq:quadratic} is a quadratic function. Even though the closed-form solution is available, we updated the solution with gradient descent steps in order to prevent undesirable jumps. The overall pseudocode to estimate domain adaptation or morphology transformation
matrix is given in Algorithm \ref{alg:mtm_algorithm}. In Figure \ref{fig:SAE_of_p_and_l_on_D_after_DA}b we show the representation errors of the transformed set of healthy signals of user $l$ on $\bm{D^p}$.

\begin{algorithm}
\caption{SR-based MTM Finding Algorithm}
\begin{algorithmic}[1]

\Procedure{Domain\_Adaptation}{$\bm{D^p}, \bm{S^l}, \gamma, \eta,$ epochs}
    \State $\bm{Q_{l \rightarrow p}} \gets \bm{I}_{N \times N}$
    \For{$i \gets 1$ to epochs}
        \State $\bm{\widehat{S}^l} \gets \bm{Q_{l \rightarrow p}} \bm{S^l}$ \Comment{Domain Adaptation}
        \State $\bm{\widehat{s}_i^l} \gets \dfrac{\bm{\widehat{s}_i^l}}{ \left\| \bm{\widehat{s}_i^l} \right\|_2}$ \Comment{Normalization}
        

        \State $\bm{X^l} \gets \underset{\bm{X^l}}{\arg\min} \left\| \bm{\widehat{S}^l} - \bm{D^p} \bm{X^l} \right\|_2^2 + \lambda \left\| \bm{X^l} \right\|_1$
        
        
        \State $\begin{aligned} \nabla \bm{Q_{l \rightarrow p}} \gets \, &\left(\left(1 + \gamma\right)\bm{Q_{l \rightarrow p}} - \gamma \bm{I}_{N \times N}\right)\bm{S^l} {\bm{S^l}}^T \\ &- \bm{D^p X^l} {\bm{S^l}}^T \end{aligned}$
        
        \State $\bm{Q_{l \rightarrow p}} \gets \bm{Q_{l \rightarrow p}} - \eta \nabla \bm{Q_{l \rightarrow p}}$
    \EndFor
    \State \Return $\bm{Q_{l \rightarrow p}}$
\EndProcedure

\end{algorithmic}
\label{alg:mtm_algorithm}
\end{algorithm}


\subsection{Ensemble Learning} \label{sec:ensemble_learning}
In Sections \ref{sec:NPE} and \ref{sec:LAE}, we explain two different methods to efficiently calculate approximation errors for ECG signals. In \cite{carrera2016ecg}, Carrera \textit{et al.} show that approximation errors can be used to classify normal and abnormal heartbeats by thresholding. However, in their work the threshold must be determined experimentally, thus they report the AUC of their method's Receiver Operating Characteristic (ROC) curve. We propose a probabilistic method to determine the classification threshold automatically from the generated training data for a specific user (the dataset generation will be explained in Section \ref{sec:training}). First, we calculate the NPE for normal and abnormal beats as described in Section \ref{sec:NPE}. Then, we estimate the parameters of two probability distributions that model the normal and abnormal errors using maximum likelihood estimation (MLE). Finally, during testing, we calculate the likelihood of the test signal's NPE belonging to one of the two distributions and classify it to the one with the higher likelihood. Instead of solely depending on this system for ECG classification, however, we train a 1-D CNN classifier for each user and consult this probabilistic classifier only when the CNN's confidence level is low. By doing so, we create an ensemble classifier.

\begin{figure*}
    \captionsetup{font=footnotesize}
    \centering
    \includegraphics[draft=false,width=\textwidth]{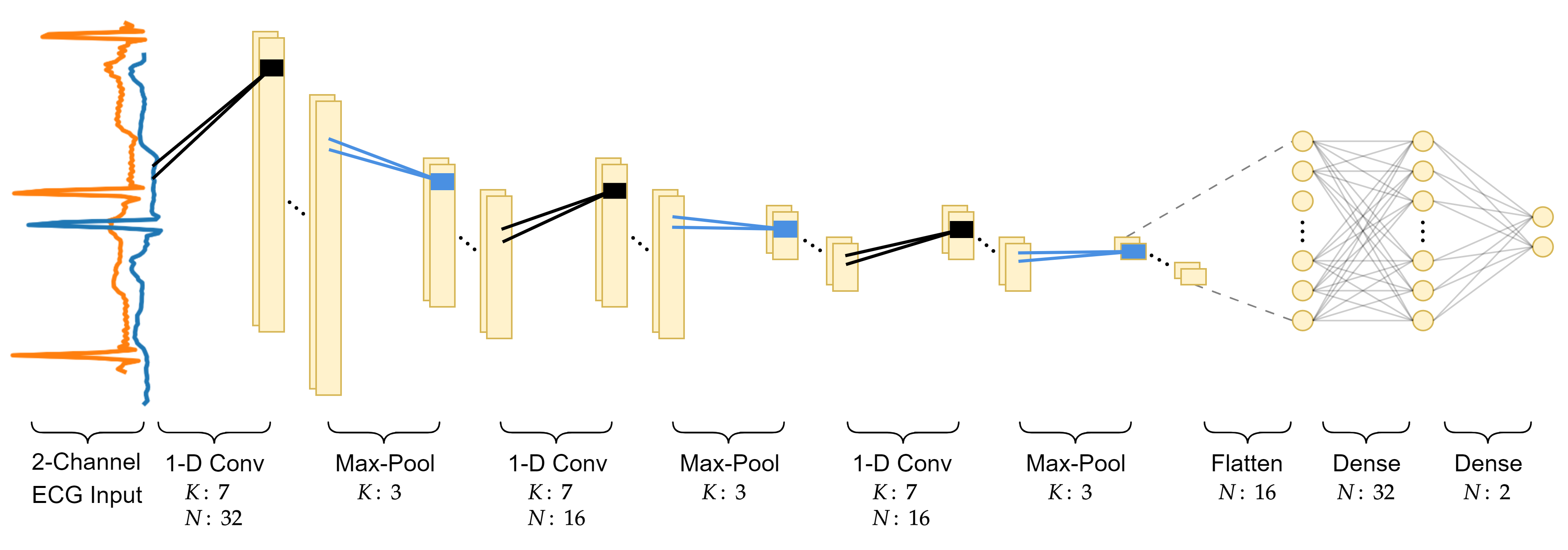}
    \caption{CNN architecture used in all experiments. K is the filter size, and N is the number of neurons in the layer.}
    \label{fig:cnn_arch}
\end{figure*}

\section{Results} \label{sec:results}
\subsection{Experimental Setup}
In our experiments, we used the ECG data from the benchmark MIT-BIH arrhythmia database \cite{moody2001impact,goldberger2000physiobank}. The database consists of two-channel ECG records from 48 different patients. Each record is approximately half an hour long with beat labels for different types of heartbeats.

\begin{figure}[!htbp]
    \captionsetup{font=footnotesize}
    \centering
    \includegraphics[draft=false,width=0.47\textwidth]{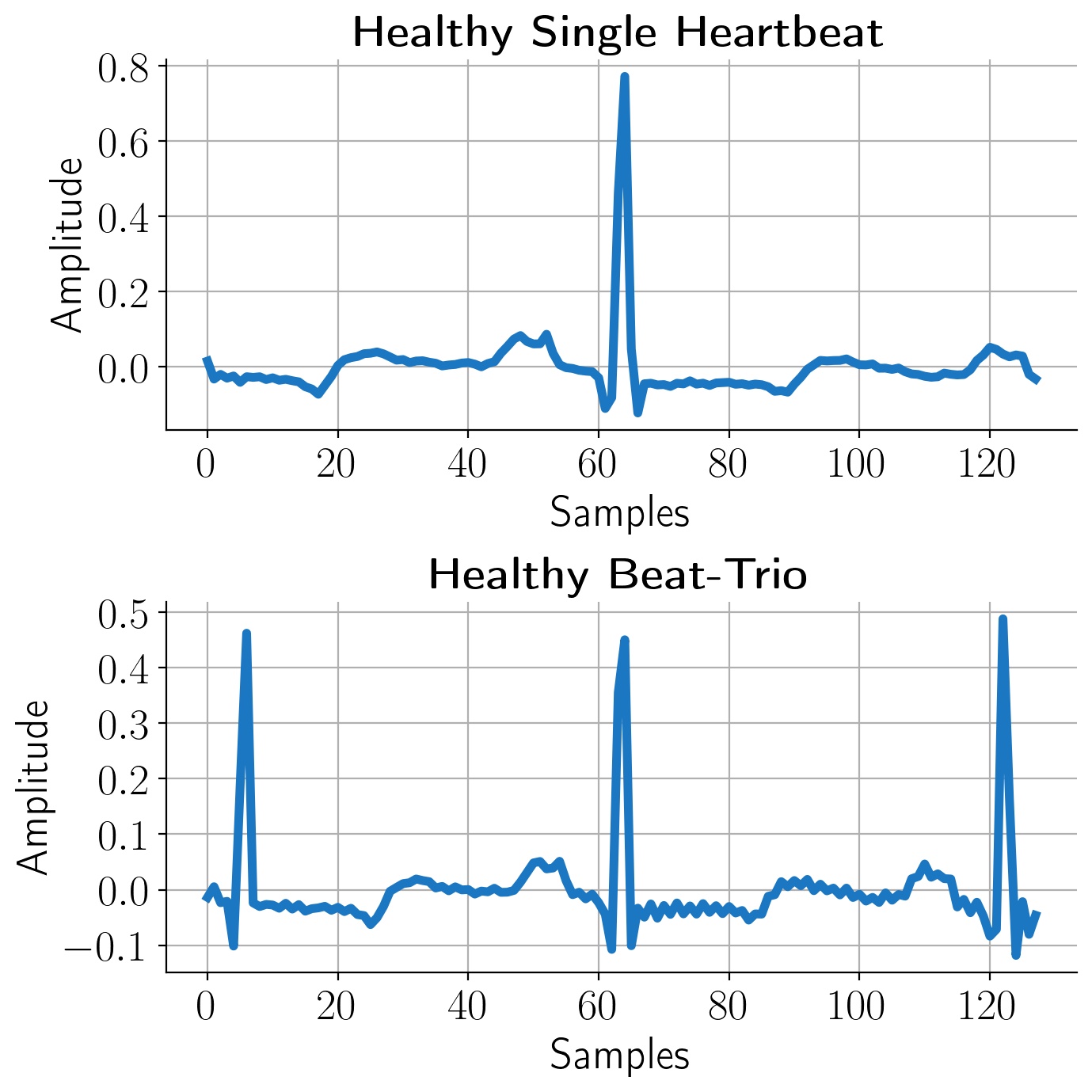}
    \caption{Single normal heartbeat and a beat-trio.}
    \label{fig:single_and_beattrio_100}
\end{figure}

We set each beat to be represented by 128 samples by resampling it. In the literature, there are two common ways of representing a beat: a single beat and a beat-trio. In both representations, the central R-peak is considered to be the reference point for the given beat. To get the single beat representation, we find the adjacent R-peaks and move 10\% inwards towards the central peak, extract that segment, and resample to 128 samples. To capture the temporal morphological characteristics of the beats, we construct a beat-trio. To get the beat-trio representation, we again find the adjacent R-peaks but now move 10\% outwards, thus the segment includes the adjacent R-peaks. Examples of single beat and beat-trio pairs are plotted in Figure \ref{fig:single_and_beattrio_100}. We use the annotations provided by the MIT-BIH database to locate any R-peaks, but for cases where peak locations are not readily available, many automated robust QRS detection algorithms have been proposed \cite{pantompkins, li1995detection, robustpeakkiranyaz}.

Similar to the experimental setup in \cite{kiranyaz2017personalized}, in this study the Association for the Advancement of Medical Instrumentation (AAMI) recommendations \cite{AAMI} are followed: AAMI categorizes heartbeat types as N (beats occurring in the sinus mode), V (ventricular ectopic beats), S (supraventricular ectopic beats), F (fusion beats) and Q (uncategorizable beats).  In this study, we considered N beats as normal beats and the other 4 types as abnormal beats. Among the 48 patient records in the MIT-BIH arrhythmia database, 34 records are used. Patients 102, 104, 107, 217 are excluded because their records come from a pacemaker. Patients 105, 114, 201, 202, 207, 209, 213, 222, 223, and 234 are excluded as they show high variations among their beats. In order to comply with the AAMI recommendation, when training an individual user's classifier, only normal heartbeats recorded within the first five minutes are included in the training data, and any abnormal beats are excluded. This small set of normal heartbeats is then combined with the datasets of the remaining 33 users according to the algorithms described in Section \ref{sec:training} (e.g., with/without domain adaptation, ABS, etc.). The abnormal beats from the first five minutes and the beats in the remaining twenty-five minutes make up that user's test data.

The performances of the proposed and competing methods are compared using the following metrics:
\begin{equation}
    \text{F1-Score} = 2 \cdot \frac{ \text{Precision} \cdot \text{Recall}}{  \text{Precision} + \text{Recall}},
\end{equation}
\begin{equation}
    \text{Specificity} = \frac{\text{TN}}{\text{TN} + \text{FP}},
\end{equation}
\begin{equation}
    \text{Precision} = \frac{\text{TP}}{\text{TP} + \text{FP}},
\end{equation}
\begin{equation}
    \text{Recall} = \frac{\text{TP}}{\text{TP} + \text{FN}},
\end{equation}
where a positive response corresponds to an abnormal beat, and true positives (TP), true negatives (TN), false positives (FP), and false negatives (FN) are computed from the predicted and ground-truth binary labels. We used the macro-average method for overall tables unless otherwise noted. In some specific cases, we analyzed user-specific performance metrics. 

\subsection{SAE vs NPE-based Anomaly Detection}
\begin{figure}[!htbp]
    \captionsetup{font=footnotesize}
    \centering
    \includegraphics[draft=false,width=0.47\textwidth]{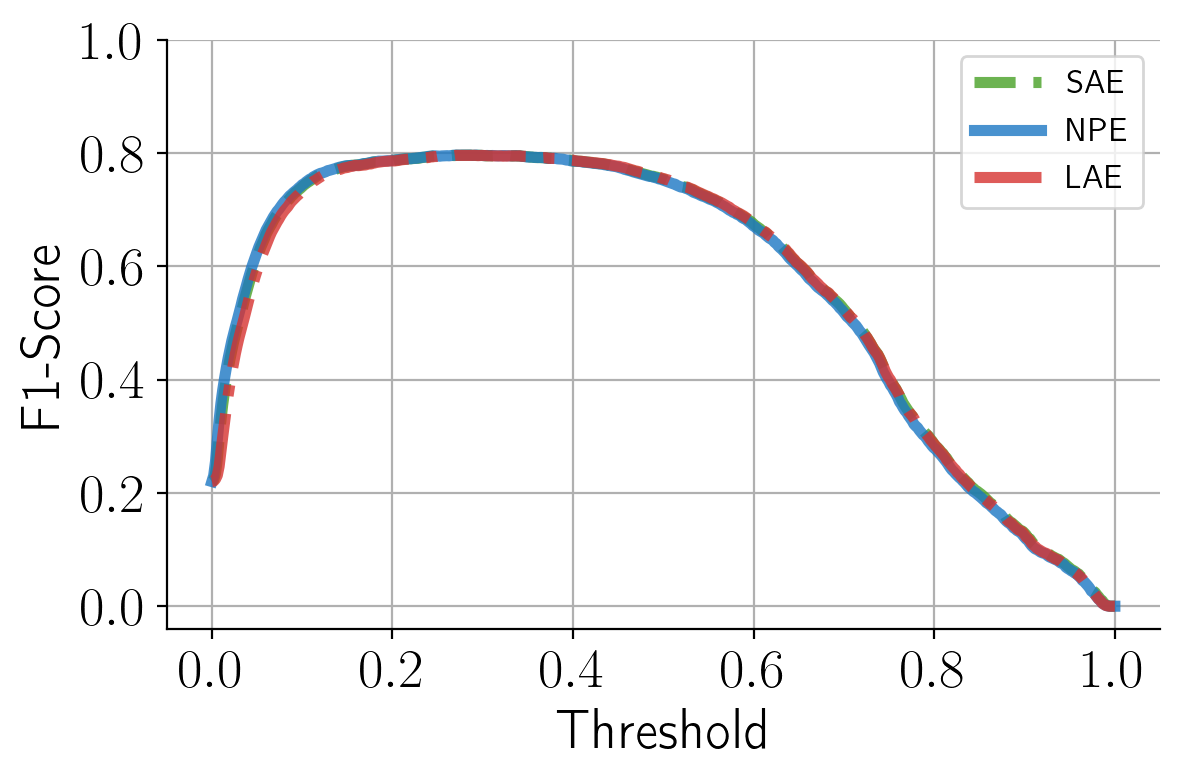}
    \caption{F1-Score with different thresholds for different representation error calculation techniques.}
    \label{fig:F1overThresholds}
\end{figure}

For our first experiment, we consider only the sparse representation-based classifier with the aforementioned methods to obtain representation errors, namely SAE, NPE, and LAE. The dictionary for each user is constructed using only the healthy beats from their first five minutes of recording. In all our experiments, we used $N=128$, $n=20$, which is experimentally observed as slightly the best setup.
In the test case, the energy of the representation error is calculated in three different ways: SAE, NPE, and LAE. Then, a predefined threshold is applied to decide whether the signal is abnormal or not. There is no question that the threshold chosen will affect the receiver operating characteristic of the classifier. As mentioned previously, ECG beats used for both training and testing are normalized to have unit energy. We can clearly see from Figure \ref{fig:F1overThresholds} that the behavior of the F1 curve is almost identical for SAE, NPE, and LAE when we draw the F1-Score for thresholds in the interval $\left[0,1\right]$.

\subsection{1-D CNN Classifier with Baseline vs ABS vs SR-based Domain Adaptation} \label{sec:training}

\begin{table*}[!htbp]
    \captionsetup{font=footnotesize}
    \centering
    {
    \setlength\tabcolsep{25pt}
    \renewcommand{\arraystretch}{1.2}
    \resizebox{\textwidth}{!}{\begin{tabular}{lccccc}
    \hline
    \rowcolor{gray!30} \textbf{Method} & \textbf{Accuracy} & \textbf{Specificity} & \textbf{Precision} & \textbf{Recall} & \textbf{F1-Score} \\
    \hline
    \multicolumn{6}{l}{\textbf{CNN}} \\
    \hline
    Kiranyaz \textit{et al.} \cite{kiranyaz2016real} $\diamond$ & 0.959 & 0.971 & 0.842 & 0.888 & 0.864 \\
    Zhai \textit{et al.} \cite{zhai2018automated} $\diamond$ & 0.968 & 0.976 & 0.879 & 0.920 & 0.899 \\
    Li \textit{et al.} \cite{li2019automated} & 0.920 & 0.918 & 0.628 & 0.933 & 0.751 \\
    \hline
    \multicolumn{6}{l}{\textbf{GAN}} \\
    \hline
    Zhou \textit{et al.} \cite{zhou2021fully} $\diamond\,*$ & 0.979 & 0.989 & 0.908 & 0.897 & 0.902 \\
    Shaker \textit{et al.} Two-stage \cite{shaker2020generalization} & 0.986 & 0.988 & 0.886 & 0.964 & 0.924 \\
    Shaker \textit{et al.} End-to-end \cite{shaker2020generalization}  & 0.987 & 0.990 & 0.901 & 0.959 & 0.929 \\
    \hline
    \multicolumn{6}{l}{\textbf{SR-based $\diamond\,*$}} \\
    \hline
    SAE-based & 0.947 & 0.968 & 0.779 & 0.794 & 0.786 \\
    NPE-based (ours) & 0.947 & 0.968 & 0.779 & 0.794 & 0.786 \\
    \hline
    \multicolumn{6}{l}{\textbf{CNN} $\diamond\,*$} \\
    \hline
    ABS \cite{kiranyaz2017personalized} & 0.977 & \textbf{0.995} & \textbf{0.956} & 0.825 & 0.886 \\
    Baseline (ours) & 0.965 & 0.987 & 0.899 & 0.809 & 0.852 \\
    Domain Adaptation (ours) & 0.978 & 0.987 & 0.911 & 0.907 & 0.909 \\
    Ensemble (ours) & \textbf{0.982} & 0.988 & 0.919 & \textbf{0.937} & \textbf{0.928} \\
    Ensemble (avg.) (ours) & 0.981 & 0.988 & 0.918 & 0.926 & 0.922 \\
    Energy-efficient (40\%) (ours) & 0.973 & 0.990 & 0.920 & 0.859 & 0.888 \\
    \hline
    \end{tabular}}
    }
    \caption{Comparison of ABS \cite{kiranyaz2017personalized}, NPE-based (ours), Baseline (ours), Domain Adaptation (ours), Ensemble Classification (ours), and Energy-efficient Classification (ours) methods. The classification performances of former studies, including global and one-shot classifiers, are presented. The results show that our personalized zero-shot ensemble classifier surpasses in F1-Score all the other methods, and is on-par with \cite{shaker2020generalization}, even though \cite{shaker2020generalization} is a global one-shot GAN-based classifier with a signal size of 300 (as opposed to 128). The confidence threshold for the ensemble classifier is chosen using the validation set. The average ensemble classifier results show the average over all possible confidence thresholds. \\
    $\diamond$ Personalized classifiers. \\
    * Zero-shot classifiers. }
    \label{tab:comparison_others}
\end{table*}

To build personalized classifiers for every user, we choose the SOTA classifier for ECG anomaly detection described in \cite{kiranyaz2017personalized} and train it with user-specific training data. Following the recommendations of AAMI, only the normal heartbeats from the first five minutes of the recording are included in that user's training data. After that, this small dataset of normal signals is combined with both normal and abnormal signals of other users using three different strategies: 

\underline{(i) Baseline Method:} The normal heartbeat dataset of user $p$ is directly combined with the dataset of other users including both normal and abnormal signals without any changes (other than a standard normalization). Specifically, all of the abnormal signals for the remaining 33 users are added along with some of their healthy signals so that the number of abnormal beats is equal to the number of normal beats.

(ii) \underline{Abnormal Beat Synthesis:} The normal heartbeat dataset of user $p$ is combined with the abnormal beat synthesis technique explained in detail in Section \ref{sec:ABS}.

(iii) \underline{Sparse Representation Based Domain Adaptation:} The proposed sparse representation-based domain adaptation technique is applied to the training dataset explained in the baseline method. As such, when creating the training dataset of user $p$, 33 different MTMs are estimated that transform the beats of each registered user $l$ to user $p$. The MTMs are applied to both normal and abnormal beats of the registered users. The results presented here use $\gamma = 0.2$ with a learning rate of $0.002$ and $25$ steps.

As a classifier, we use a 1-D CNN with $3$ convolutional layers followed by $2$ fully-connected layers. The convolutional layers have a kernel size of 7, no padding, and a stride of 1. Each convolutional layer is followed by a max-pooling layer with a stride of 3, and a hyperbolic tangent activation function. The first fully-connected layer has a rectified linear unit activation function, and the final layer has a log-softmax activation. The number of neurons per layer is 32, 16, 16, and 32 respectively and the final (output) layer has 2 neurons, one for each class. A diagram of the network structure can be found in Figure \ref{fig:cnn_arch}. We used the same network structure for ABS and the proposed domain adaptation based training data generator to make a fair comparison. The training and assessment of the model are performed as follows. First, the datasets are generated per patient for both single and beat-trio beat representations, using the baseline method and our domain adaptation method. These datasets are then split into training and validation datasets with an 80\% training ratio. Then, single beats and beat-trios are fed together as two-channel inputs to the CNN. We used cross-entropy loss and the weight-decayed Adam optimizer \cite{adamW} to update the weights. The model keeps training until 15 epochs have passed without improving the validation loss. The weights that achieve the lowest validation loss are then used for evaluation. We average our results over 10 independent training runs to minimize the effect of the randomly initialized parameters.

\begin{figure}[!htbp]
    \captionsetup{font=footnotesize}
    \centering
    \includegraphics[draft=false,width=0.47\textwidth]{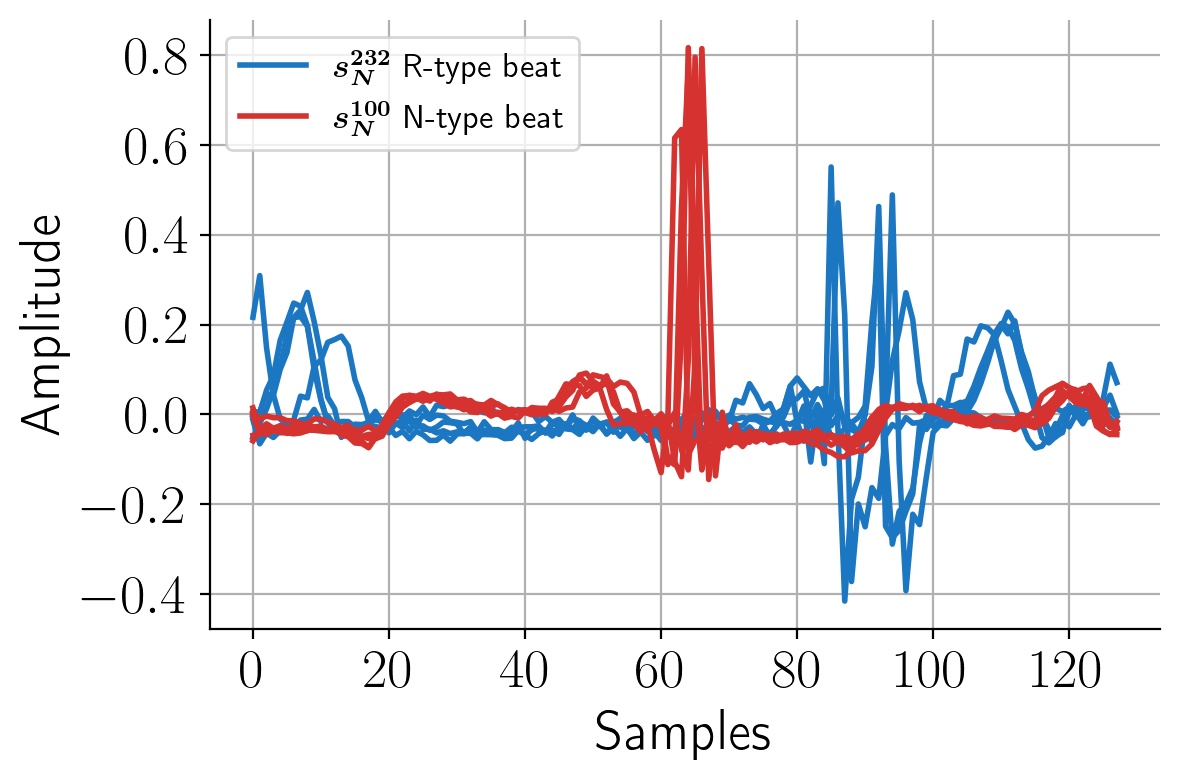}
    \caption{A set of R-type healthy beats and N-type healthy beats from patients 232 and 100, respectively. This set of R-type beats has distinct morphological differences from N-type beats.}
    \label{fig:healthy_232_vs_100}
\end{figure}

As it can be seen in Tables \ref{tab:comparison_others} and \ref{tab:cm_baseline_DA} the proposed domain adaptation method vastly improves the baseline approach and surpasses the abnormal beat generation strategy in terms of recall and F1-Score. We show that domain adaptation can greatly enhance arrhythmia detection, resulting in a fewer number of missed arrhythmia beats. Moreover, domain adaptation is especially effective for patients with normal beats that are dissimilar to that of other patients' normal beats. The AAMI standard classifies normal beats (N), left bundle branch block beats (L), right bundle branch block beats (R), atrial escape beats (e), and nodal (junctional) escape beats (j) as normal beats. However, those beats may show variations among them, to the extent that the normal beats of user $p$ are very distinct from the normal beats of user $l$, as shown in Figure \ref{fig:healthy_232_vs_100}.

\begin{table}[!htbp]
    \captionsetup{font=footnotesize}
    \centering
    {
    \setlength\tabcolsep{3pt}
    \renewcommand{\arraystretch}{1.5}
    \begin{tabular}{|ccc|c|c|c|c|c|c|}
    \hhline{~~~------}
    \multicolumn{3}{c}{} & \multicolumn{6}{c}{\cellcolor{gray!30}\textbf{Ground Truth}} \\
    \hhline{~~~------}
    \multicolumn{1}{c}{} \\[-4.5mm]
    \hhline{~~~--~~--}
    \cellcolor{gray!30} & \multicolumn{2}{|c|}{} & A & N & \multicolumn{1}{c}{} & \multicolumn{1}{c|}{} & A & N \\
    \cline{3-5} \cline{7-9}
    \cellcolor{gray!30} & \multicolumn{1}{|c|}{} & \multicolumn{1}{c|}{A} & \,\,3361\,\, & \,\,\,\,185\,\,\,\, & & A & \,\,9054\,\, & \,\,\,\,218\,\,\,\, \\
    \cline{3-5} \cline{7-9}
    \multirow{-3}{*}{\cellcolor{gray!30}\,\,\rotatebox[origin=c]{90}{\textbf{Predicted}}\,\,} & \multicolumn{1}{|c|}{} & \multicolumn{1}{c|}{\,\,N\,\,} & 10449 & 2965 & & \,\,N\,\, & 4756 & 2932 \\
    \cline{3-5} \cline{7-9}
    \end{tabular}
    }
    \caption{Confusion matrices for patient 232, obtained with the Baseline (left) and Domain Adaptation (right) methods, accumulated over 10 independent training runs. A = Abnormal, N = Normal.}
    \label{tab:cm_232_baseline_vs_DA}
\end{table}

\begin{table}[!htbp]
    \captionsetup{font=footnotesize}
    \centering
    {
    \setlength\tabcolsep{3pt}
    \renewcommand{\arraystretch}{1.5}
    \begin{tabular}{|ccc|c|c|c|c|c|c|}
    \hhline{~~~------}
    \multicolumn{3}{c}{} & \multicolumn{6}{c}{\cellcolor{gray!30}\textbf{Ground Truth}} \\
    \hhline{~~~------}
    \multicolumn{1}{c}{} \\[-4.5mm]
    \hhline{~~~--~~--}
    \cellcolor{gray!30} & \multicolumn{2}{|c|}{} & A & N & \multicolumn{1}{c}{} & \multicolumn{1}{c|}{} & A & N \\
    \cline{3-5} \cline{7-9}
    \cellcolor{gray!30} & \multicolumn{1}{|c|}{} & \multicolumn{1}{c|}{A} & \,\,64447\,\, & 7210 & & A & \,\,72212\,\, & 7065 \\
    \cline{3-5} \cline{7-9}
    \multirow{-3}{*}{\cellcolor{gray!30}\,\,\rotatebox[origin=c]{90}{\textbf{Predicted}}\,\,} & \multicolumn{1}{|c|}{} & \multicolumn{1}{c|}{\,\,N\,\,} & 15183 & 557070 & & \,\,N\,\, & 7418 & 557215 \\
    \cline{3-5} \cline{7-9}
    \end{tabular}
    }
    \caption{Confusion matrices of the Baseline (left) and Domain Adaptation (right) methods, accumulated over 10 independent training runs. A = Abnormal, N = Normal.}
    \label{tab:cm_baseline_DA}
\end{table}

\subsection{Ensemble Classification}
\label{sec:ensemble_classification}
After the CNN is trained, we estimate the parameters of the probabilistic model as described in Section \ref{sec:ensemble_learning}. For all of our experiments, we model the normal NPEs with an exponential distribution, and abnormal NPEs with a Gaussian distribution. The probability density function (PDF) of the exponential distribution is

\begin{equation}
\label{eq:expon_pdf}
f\left( x; \beta \right) = \begin{cases}
\frac{1}{\beta} e^{-x/\beta} & x \geq 0 \\
0 & x < 0
\end{cases}.
\end{equation}
The maximum likelihood estimate for the parameter $\beta$ is 

\begin{equation}
\widehat{\beta} = \underset{\beta}{\arg\max} \prod_{i=1}^{n} \frac{1}{\beta} e^{-x_i / \beta} = \dfrac{\sum_{i=1}^{n}x_i}{n},
\end{equation}
which is the sample mean. Similarly, the PDF of the Gaussian distribution is

\begin{equation}
\label{eq:gaus_pdf}
f\left( x; \mu, \sigma \right) = \frac{1}{\sigma \sqrt{2 \pi}} e^{-\frac{1}{2} \left( \frac{x - \mu}{\sigma} \right)^{2}},
\end{equation}
and the maximum likelihood estimate for the parameters $\mu$ and $\sigma$ are

\begin{equation}
    \widehat{\mu} = \dfrac{1}{n} \sum_{i=1}^{n}x_i, \quad \widehat{\sigma} = \sqrt{\dfrac{1}{n} \sum_{i=1}^{n} \left(x_i - \mu\right)},
\end{equation}
which are the sample mean and sample standard deviation respectively.

\begin{figure}[!htbp]
    \captionsetup{font=footnotesize}
    \centering
    \includegraphics[draft=false,width=0.5\textwidth]{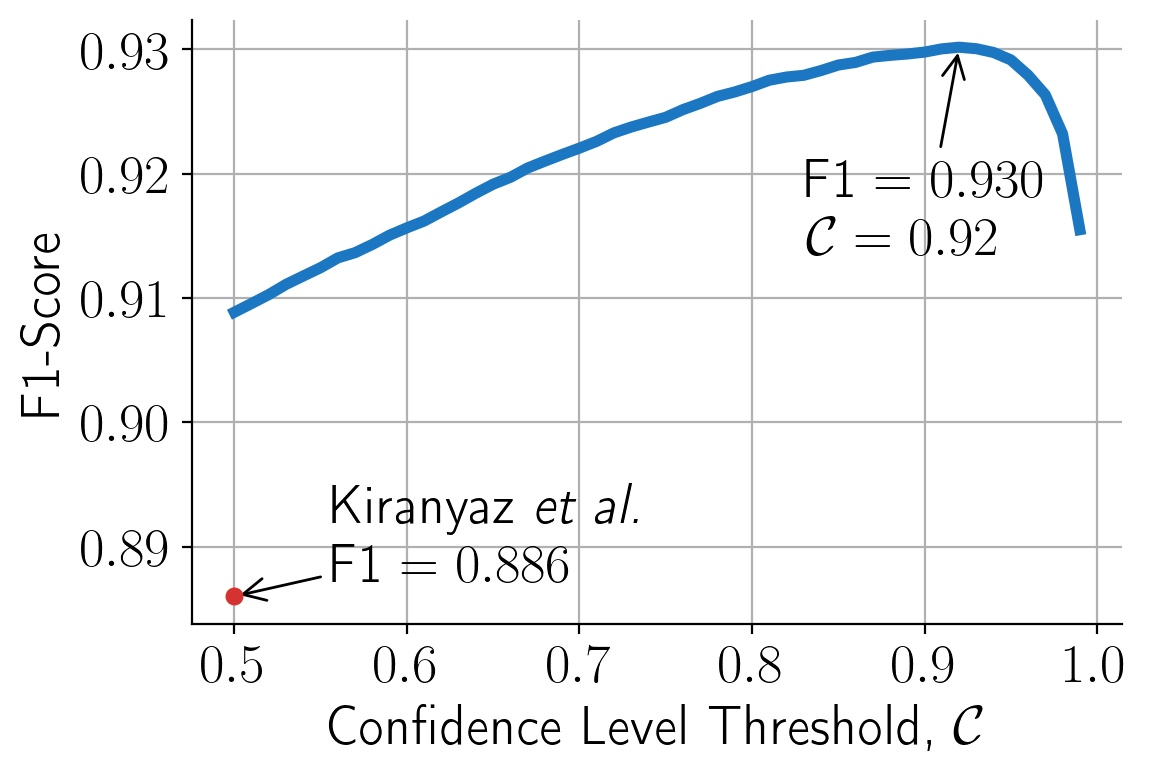}
    \caption{F1-Score for all possible values of confidence level threshold. Values too close or equal to 1 force the system to use the simple representation error-based classifier, thus achieving lower F1-Scores.}
    \label{fig:C_vs_F1}
\end{figure}

At test time, the CNN output is passed through the softmax function instead of log-softmax. Then, we choose the greater neuron output as the confidence of the CNN. If the confidence is above some threshold, $\mathcal{C}$, then the network output is used for classification. Otherwise, the NPE of the test sample is calculated and the probabilistic model is used for classification. If $\mathcal{C} \leq 0.5$, then only the CNN output is used, and if $\mathcal{C} \geq 1$, only the probabilistic model is used. We choose $\mathcal{C}$ as the confidence that maximizes F1-Score in the validation set. In the case of equally performing confidences, we choose the greater confidence. In Figure \ref{fig:C_vs_F1}, we show that any choice of $\mathcal{C}$ in a broad range results in a performance improvement. 

In Table \ref{tab:comparison_others}, a comprehensive analysis on the comparison of different algorithms is done. Among competing algorithms, only ABS \cite{kiranyaz2017personalized}, SAE \cite{carrera2016ecg}, a GAN-based ECG generation study \cite{zhou2021fully}, and the variants of the proposed algorithms are personalized zero-shot methods. The comparison is made among only personalized zero-shot methods which are marked with $\diamond\,*$. In addition to personalized zero-shot methods, the table includes a few SOTA CNN-based algorithms \cite{kiranyaz2016real, zhai2018automated, li2019automated}, as well as a SOTA GAN-based technology \cite{shaker2020generalization}. Among them, CNN-based algorithms are trained person-specific or globally, and GAN-based method is one of the aforementioned SOTA algorithms that tackle data imbalance problems. Those algorithms rely on the abnormal dataset of a given patient, partly or completely. Therefore, they cannot be directly compared to the personalized zero-shot methods.

The baseline method can be considered as the one where a person-specific network is trained in a personalized manner by using the global dataset excluding the abnormal data of the person of interest. Due to the variation in hearth beat morphology across users, such a vanilla approach has limited performance, as can be seen in Table \ref{tab:comparison_others}. This is especially true for users whose healthy (normal) heartbeat morphology differs significantly from the majority of users. We examine patient 232 from the MIT-BIH dataset as an example. The results presented in Table \ref{tab:cm_232_baseline_vs_DA} shows that the proposed algorithm can significantly improve the performance of personalized zero-shot classification performance.

Moreover the results presented in Table \ref{tab:comparison_others} clearly indicate that our ensemble classifier greatly improves our domain adaptation method and achieves the best accuracy, recall, and F1-Score among all methods, improving recall by over $11\%$ and F1-Score by over $4\%$ compared to ABS \cite{kiranyaz2017personalized}. The overall confusion matrix associated with the proposed solution can be found in Table \ref{tab:cm_consult}. As mentioned in Section \ref{sec:ensemble_classification}, the 'confidence level' threshold determines which classifier is to be used. This threshold is determined from the validation set in a completely zero-shot manner over the generated abnormal beats. However, any choice of this threshold makes the proposed approach outperform all competing algorithms as shown in Figure \ref{fig:C_vs_F1}. In Table \ref{tab:comparison_others} the average performances for 50 different thresholds selected uniformly in the interval $[0.50, 0.99]$ are presented.

\begin{table}[!htbp]
    \captionsetup{font=footnotesize}
    \centering
    {
    \setlength\tabcolsep{3pt}
    \renewcommand{\arraystretch}{1.5}
    \begin{tabular}{|ccc|c|c|c|c|c|c|}
    \hhline{~~~------}
    \multicolumn{3}{c}{} & \multicolumn{6}{c}{\cellcolor{gray!30}\textbf{Ground Truth}} \\
    \hhline{~~~------}
    \multicolumn{1}{c}{} \\[-4.5mm]
    \hhline{~~~--~~--}
    \cellcolor{gray!30} & \multicolumn{2}{|c|}{} & A & N & \multicolumn{1}{c}{} & \multicolumn{1}{c|}{} & A & N \\
    \cline{3-5} \cline{7-9}
    \cellcolor{gray!30} & \multicolumn{1}{|c|}{} & \multicolumn{1}{c|}{A} & \,\,73759\,\, & 6577 & & A & \,\,74618\,\, & 6590 \\
    \cline{3-5} \cline{7-9}
    \multirow{-3}{*}{\cellcolor{gray!30}\,\,\rotatebox[origin=c]{90}{\textbf{Predicted}}\,\,} & \multicolumn{1}{|c|}{} & \multicolumn{1}{c|}{\,\,N\,\,} & 5871 & 557703 & & \,\,N\,\, & 5012 & 557690 \\
    \cline{3-5} \cline{7-9}
    \end{tabular}
    }
    \caption{Confusion matrices of the Ensemble Classifier averaged over all $\mathcal{C}$ (left) and $\mathcal{C}$ chosen from the validation set (right), accumulated over 10 independent training runs. A = Abnormal, N = Normal.}
    \label{tab:cm_consult}
\end{table}

\subsection{Ablation Studies}
\label{sec:ablation}
We take a moment to discuss the empirical analysis conducted to select the hyper-parameters of the proposed methodology.

\subsubsection{Model Selection}
As we mention in Section \ref{sec:training}, we use the compact CNN model from the landmark work \cite{kiranyaz2017personalized}. The reason to use such a compact model is 3-fold:
\begin{itemize}
    \item[i.] We want to ensure a fair comparison with \cite{kiranyaz2017personalized}.
    \item[ii.] The model is parameter efficient with only 6K parameters and can be deployed even on mobile devices due to its memory and runtime efficiency. The model takes up only 24 KB of space and performs 350K FLOPs per input.
    \item[iii.] The nature of the patient-specific and zero-shot classification problem implies the lack of training data. This is in contrast to an abundance of data in other deep learning problems such as natural language processing or image generation. In our case, there is no need for a deeper and more complex model; on the contrary, such a deeper model has the potential to overfit the limited train data and reduce the overall performance.
\end{itemize}
Nonetheless, we train a Transformer-based model that follows a similar architecture to our CNN, but employs the attention modules proposed in the pioneering art \cite{vaswani2017attention}. This model is five times the size of our CNN and performs 7.4M FLOPs per input, yet the training performance is equal to our CNN at best and 1 F1-Score behind on certain training runs at worst.

\subsubsection{Number of Dictionary Atoms}
The number of dictionary atoms, $n$, is set to $20$, the same as the prior art \cite{carrera2016ecg} to ensure a fair comparison. Even so, it is important to discuss the effect of different numbers of atoms on the representation power of the dictionary. If $n$ is too small, the dictionary cannot represent even the small set of healthy beats of the user. Thus, NPE will be high for both healthy and abnormal beats, and we cannot perform error-based classification. Furthermore, Algorithm \ref{alg:mtm_algorithm} will fail as it depends on the representation capabilities of $\bm{D}$ to find a suitable transformation from patient $l$ to patient $p$. If $n$ is too large, the dictionary constructed over the normal beats may be able to represent even the abnormal beats. Thus the choice of $n$ will depend on the number of healthy beats used to learn the dictionary and the variations within the healthy beats.

\subsubsection{Loss Hyper-parameters}
The hyper-parameters in all of our lost functions are empirically selected, and the best-performing parameter setup is reported. Throughout all of the $\ell_1$ regularization terms, $\lambda$ is chosen as 0.01. In Eq. \ref{eq:domain_adaptation}, $\gamma$ is chosen as 0.2.

\section{A practical ECG monitoring scheme with energy-saving mechanism}
\label{sec:energy_eff}
Continuous cardiac monitoring on wearable devices has the potential to increase early diagnosis, provide personalized care, and reduce the risk of sudden cardiovascular issues. However, on-device neural network-based solutions or wireless transmission for off-device arrhythmia detection may be energy inefficient to be continuously employed on mobile wear \cite{carrera2016ecg, mamaghanian2011compressed}. Thus, in this section, we describe an energy-preserving ECG monitoring scheme for using on wearable devices. 

\begin{figure}[!htbp]
    \captionsetup{font=footnotesize}
    \centering
    \includegraphics[draft=false,width=0.47\textwidth]{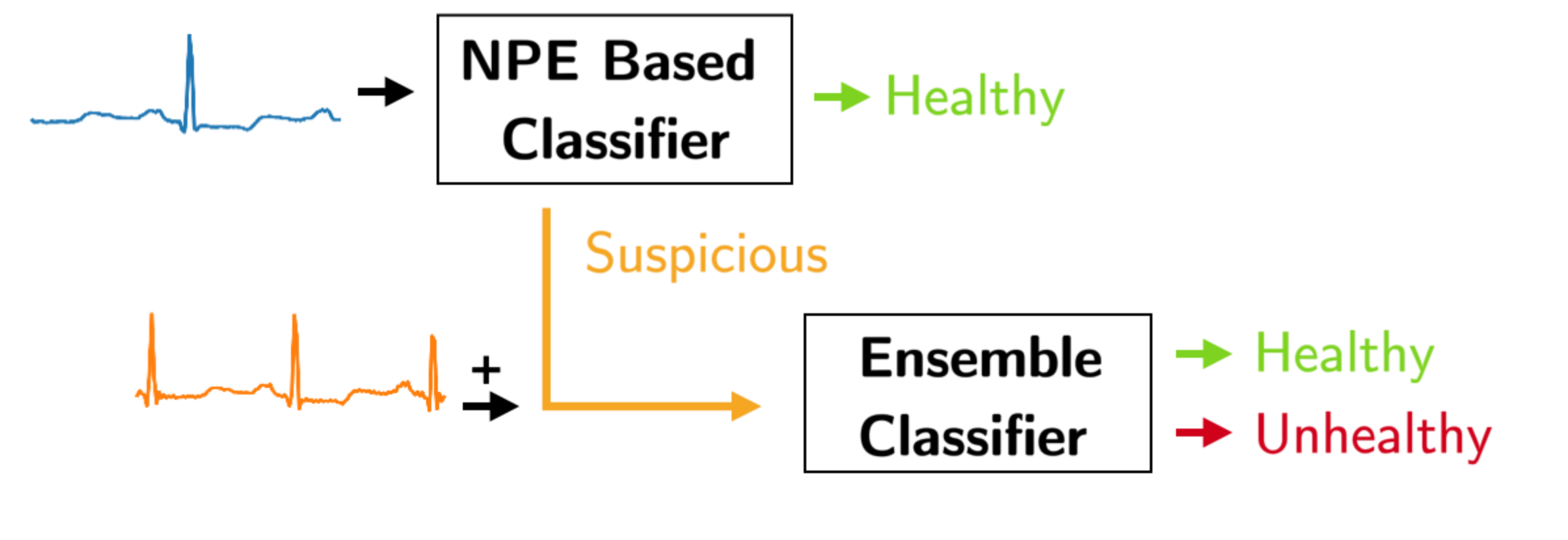}
    \caption{Overall system model of the energy-saving ECG monitoring scheme.}
    \label{fig:practical_energy_saving}
\end{figure}

The idea is that a significant number of normal beats can be classified with very high confidence using simple and energy-efficient NPE or LAE techniques. In this manner in a practical continuous monitoring system, beats can be first classified with NPE and the suspicious ones can be filtered out to be analyzed through the proposed approach. The proposed energy-saving ECG monitoring scheme is illustrated Figure \ref{fig:practical_energy_saving}.

\begin{figure}[!htbp]
    \captionsetup{font=footnotesize}
    \centering
    \includegraphics[draft=false,width=0.5\textwidth]{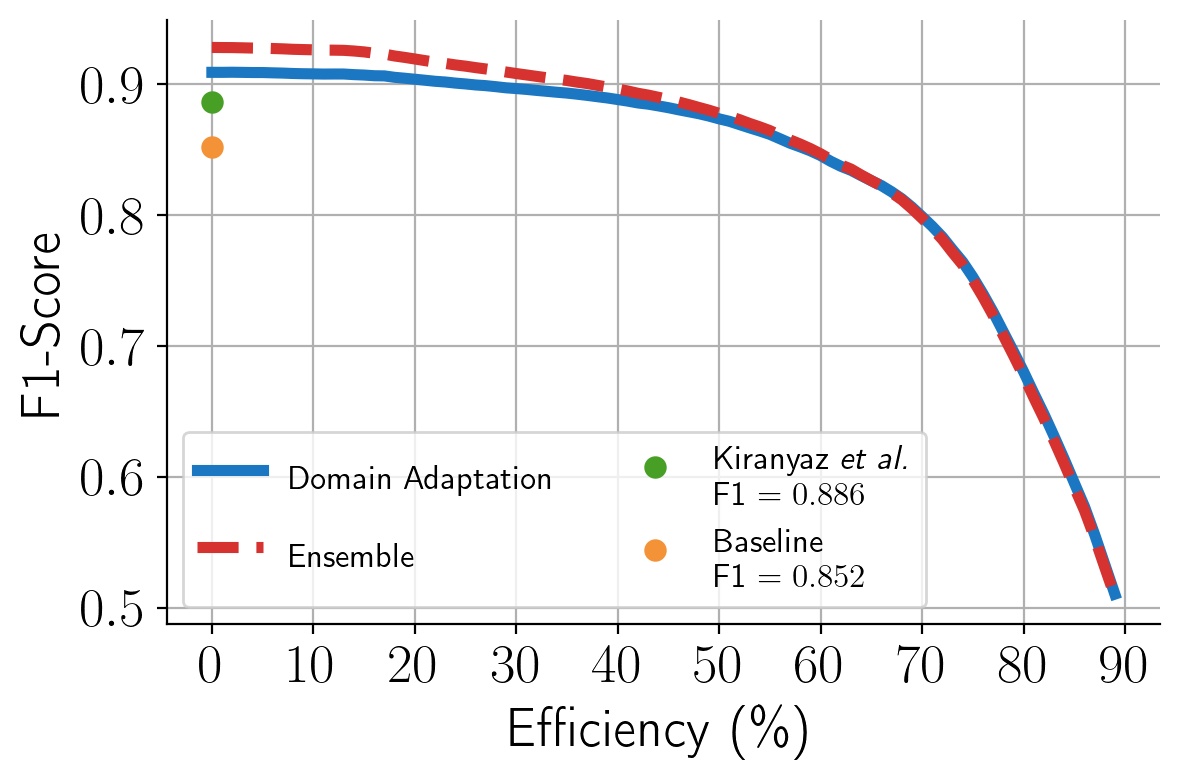}
    \caption{Energy efficiency over F1-Scores, where the x-axis shows the percentage of test samples that are classified solely based on NPE, hence being more computationally efficient.}
    \label{fig:Efficiency_vs_F1}
\end{figure}

Figure \ref{fig:Efficiency_vs_F1} shows that up to 40\% of all the 64391 samples in the test dataset can be classified solely based on NPE without sacrificing significantly from the F1 performance.

\section{Discussion}
\label{sec:discussion}
\subsection{How Does Representation Based Anomaly Detection Differ from Representation Based Classifiers (Sparse and Collaborative Representations)?}
The conventional representation-based (dictionary-based) classification schemes include a pre-established linear representation dictionary that contains sub-dictionaries that each store representation atoms that correspond to a specific class. 
After a test query sample has been received, the representation coefficient $\bm{x}$ is estimated either by $\ell_1$-minimization or by the regularized least-squares. If it is estimated with sparse recovery algorithms such as $\ell_1$-minimization, the method is referred to as sparse representation-based classification (SRC \cite{SRC1} ); likewise, if it is estimated by the least-squares solution, the method is called collaborative representation (CRC \cite{collaborative}). 
Indeed, both SRC and CRC are collaborative representation methods since the representation coefficient vector is calculated over an overall dictionary instead of using class-specific sub-dictionaries \cite{collaborative}. The problem of anomaly detection without the abnormal data during the training phase is a one-class classification problem, in which the dictionary contains representative atoms for one class only, therefore it cannot be categorized as a collaborative representation.

\subsection{Limitations and Future Work}
This paper provides a complete pipeline of continuous ECG monitoring for the following realistic scenario: A new user is registered to the (monitoring) system, and only a few minutes of normal (healthy) beats are acquired for continuous monitoring afterwards. 
The pipeline consists of person-specific calibration of the algorithm and a strategy to improve the battery life span by only choosing the suspicious beats to be analyzed using the proposed approach. 
The personalized zero-shot anomaly detection was considered only in a few prior works, and our sparse representation-based domain adaption significantly improves the SOTA performance. For a fair comparison, we use the same network structure that was used in the competing SOTA algorithm. In that way, we proved that the proposed realistic data generation technique is highly superior. As a near-feature plan, more advanced network structures will be investigated.
For the first time in this study, we provide a complete scheme with energy-efficient pre-filtering of the suspicious beats. We use the left null space basis as a simple projection matrix and achieve up to $\%40$ reduction in the overall complexity.
Therefore, this serves as the pilot study; however, we believe that there is still room to improve both the performance and the computational efficiency. For this purpose, we will be investigating more advanced projection matrices.





\section{Conclusion}
\label{sec:conclusion}
This paper addresses the problem of personalized and zero-shot abnormal heartbeat detection through sparse coding, sparse representation-based domain adaptation, and a robust and energy-efficient ensemble classifier that achieves state-of-the-art performance on the MIT-BIH Arrhythmia Database. We draw the following conclusive remarks on the proposed methodology based on the rigorous experiments outlined in Section \ref{sec:results}.

\underline{Method \ref{sec:NPE} (and \ref{sec:LAE})}: By constructing the left annihilator matrix $\bm{F^p}$ of the dictionary $\bm{D^p}$ for user $\bm{p}$, instead of iteratively solving a sparse recovery problem, we can directly project the error component of a given test signal $\bm{s_i^p}$ onto the null space of $\bm{D^p}$. We show that classification based on the representation error component, $\widetilde{\bm{e}}$, provides significant computational improvements (up to 20 times) without any performance drawbacks. We provide an alternative solution in \ref{sec:LAE} that achieves the same performance but may be computationally more suitable for different hardware configurations. We provide the computational analysis in Tables \ref{tab:FLOPs} and \ref{tab:runtimes}. We conclude that classification from representation error is the most energy efficient, but as a simple thresholding classifier, it eventually yields the worst performance.

\underline{Method \ref{sec:SR-DA}}: Using our novel sparse representation-based domain adaptation method and morphology transformation matrices described in Algorithm \ref{alg:mtm_algorithm}, it is possible to perform inter-patient beat transfer with adapted ECG morphology. Thus, large training sets can be created for each user by transferring original healthy and abnormal beats from all of the remaining users. Our CNN classifier achieves an intermediary state-of-the-art performance when trained on these datasets, resulting in $97.8\%$ accuracy and $90.9\%$ F1-Score. Through domain adaptation, we eliminate the need to generate synthetic abnormal beats and drastically improve personalized and zero-shot abnormal beat classification performance.

\underline{Method \ref{sec:ensemble_learning}}: We fuse methods \ref{sec:NPE} and \ref{sec:SR-DA} together to create an ensemble classifier. The CNN classifier consults the simple error-based classifier based on a confidence threshold. In Figure \ref{fig:C_vs_F1}, we show that the ensemble classifier boosts the results for any choice of confidence threshold. Nonetheless, we select the confidence threshold through the validation set and achieve the state-of-the-art performance with $98.2\%$ accuracy and $92.8\%$ F1-Score. We present the performance levels achieved by our method and the prior art in Table \ref{tab:comparison_others} for comparative evaluation.

\underline{Monitoring System (\ref{sec:energy_eff})}: Finally, we build an energy-efficient monitoring system by utilizing the computational efficiency of Method \ref{sec:NPE} and superior performance of Method \ref{sec:ensemble_learning}. The basis for such a system is that a large number of healthy beats will be well-represented in the dictionary and can be classified through NPE. An outline of the system is provided in Figure \ref{fig:practical_energy_saving}. The monitoring system can classify up to $40\%$ of all test samples solely based on the NPE (Method \ref{sec:NPE}) without a noticeable sacrifice in performance. The suspicious beats with high NPE are filtered out and analyzed further by the ensemble classifier (Method \ref{sec:ensemble_learning}). We conclude that there is potential for energy-efficient monitoring through representation-based classification suitable for wearable devices.

\ifCLASSOPTIONcaptionsoff
  \newpage
\fi

\bibliographystyle{IEEEtran}
\bibliography{references}
\end{document}

%% file: figures_tikz/linear_degrading_system.tex
\begin{figure}[H]
    \captionsetup{font=footnotesize}
\centering

\tikzset{every picture/.style={line width=0.75pt}} 

\begin{tikzpicture}[x=0.60pt,y=0.75pt,yscale=-1,xscale=1]

\draw [color={rgb, 255:red, 65; green, 117; blue, 5 }  ,draw opacity=1 ][line width=1.5]    (45,90) -- (165,90) ;
\draw  [color={rgb, 255:red, 65; green, 117; blue, 5 }  ,draw opacity=1 ][line width=1.5]  (65,30) -- (80,90) -- (50,90) -- cycle ;
\draw  [color={rgb, 255:red, 65; green, 117; blue, 5 }  ,draw opacity=1 ][line width=1.5]  (105,30) -- (120,90) -- (90,90) -- cycle ;
\draw  [color={rgb, 255:red, 65; green, 117; blue, 5 }  ,draw opacity=1 ][line width=1.5]  (145,30) -- (160,90) -- (130,90) -- cycle ;
\draw [color={rgb, 255:red, 65; green, 117; blue, 5 }  ,draw opacity=1 ][line width=1.5]    (294,89) -- (415,90) ;
\draw  [color={rgb, 255:red, 65; green, 117; blue, 5 }  ,draw opacity=1 ][line width=1.5]  (314,29) -- (329,89) -- (299,89) -- cycle ;
\draw  [color={rgb, 255:red, 208; green, 2; blue, 27 }  ,draw opacity=1 ][line width=1.5]  (354,29) -- (369,89) -- (339,89) -- cycle ;
\draw  [color={rgb, 255:red, 65; green, 117; blue, 5 }  ,draw opacity=1 ][line width=1.5]  (394,29) -- (409,89) -- (379,89) -- cycle ;
\draw  [color={rgb, 255:red, 0; green, 0; blue, 0 }  ,draw opacity=1 ][fill={rgb, 255:red, 0; green, 0; blue, 0 }  ,fill opacity=1 ] (189,10) -- (269,10) -- (269,110) -- (189,110) -- cycle ;
\draw  [line width=1.5]  (220,140.6) -- (225.44,140.6) -- (225.44,120) -- (234.56,120) -- (234.56,140.6) -- (240,140.6) -- (230,150) -- cycle ;
\draw [color={rgb, 255:red, 126; green, 211; blue, 33 }  ,draw opacity=1 ][line width=1.5]    (45,239) -- (165,239) ;
\draw [color={rgb, 255:red, 126; green, 211; blue, 33 }  ,draw opacity=1 ][line width=1.5]    (295,240) -- (415,240) ;
\draw  [color={rgb, 255:red, 208; green, 2; blue, 27 }  ,draw opacity=1 ][line width=1.5]  (354,29) .. controls (354.15,29) and (354.27,29.06) .. (354.36,29.18) .. controls (354.46,29.3) and (354.47,29.45) .. (354.42,29.61) .. controls (354.35,29.8) and (354.21,29.95) .. (354,30.06) .. controls (353.76,30.19) and (353.48,30.25) .. (353.17,30.23) .. controls (352.81,30.2) and (352.49,30.09) .. (352.2,29.89) .. controls (351.87,29.66) and (351.64,29.36) .. (351.5,29) .. controls (351.35,28.6) and (351.34,28.19) .. (351.47,27.76) .. controls (351.62,27.3) and (351.91,26.89) .. (352.33,26.55) .. controls (352.8,26.17) and (353.35,25.93) .. (354,25.81) .. controls (354.69,25.69) and (355.39,25.73) .. (356.08,25.93) .. controls (356.82,26.15) and (357.45,26.52) .. (357.97,27.05) .. controls (358.52,27.61) and (358.86,28.26) .. (359,29) .. controls (359.14,29.78) and (359.05,30.55) .. (358.69,31.3) .. controls (358.32,32.09) and (357.73,32.76) .. (356.92,33.29) .. controls (356.07,33.85) and (355.09,34.19) .. (354,34.31) .. controls (352.86,34.44) and (351.75,34.3) .. (350.67,33.91) .. controls (349.54,33.5) and (348.61,32.87) .. (347.87,32.01) .. controls (347.09,31.12) and (346.64,30.12) .. (346.5,29) .. controls (346.35,27.84) and (346.56,26.72) .. (347.14,25.64) .. controls (347.74,24.51) and (348.64,23.59) .. (349.83,22.87) .. controls (351.07,22.12) and (352.46,21.68) .. (354,21.56) .. controls (355.59,21.44) and (357.12,21.67) .. (358.58,22.25) .. controls (360.09,22.85) and (361.33,23.74) .. (362.3,24.93) .. controls (363.29,26.14) and (363.86,27.5) .. (364,29) .. controls (364,29) and (364,29) .. (364,29) ;
\draw  [color={rgb, 255:red, 126; green, 211; blue, 33 }  ,draw opacity=1 ][line width=1.5]  (50,186) .. controls (53.31,186) and (56,183.31) .. (56,180) -- (74,180) .. controls (74,183.31) and (76.69,186) .. (80,186) -- (80,233) .. controls (76.69,233) and (74,235.69) .. (74,239) -- (56,239) .. controls (56,235.69) and (53.31,233) .. (50,233) -- cycle ;
\draw  [color={rgb, 255:red, 126; green, 211; blue, 33 }  ,draw opacity=1 ][line width=1.5]  (91,186) .. controls (94.31,186) and (97,183.31) .. (97,180) -- (115,180) .. controls (115,183.31) and (117.69,186) .. (121,186) -- (121,233) .. controls (117.69,233) and (115,235.69) .. (115,239) -- (97,239) .. controls (97,235.69) and (94.31,233) .. (91,233) -- cycle ;
\draw  [color={rgb, 255:red, 126; green, 211; blue, 33 }  ,draw opacity=1 ][line width=1.5]  (130,186) .. controls (133.31,186) and (136,183.31) .. (136,180) -- (154,180) .. controls (154,183.31) and (156.69,186) .. (160,186) -- (160,233) .. controls (156.69,233) and (154,235.69) .. (154,239) -- (136,239) .. controls (136,235.69) and (133.31,233) .. (130,233) -- cycle ;
\draw  [color={rgb, 255:red, 126; green, 211; blue, 33 }  ,draw opacity=1 ][line width=1.5]  (299,186) .. controls (302.31,186) and (305,183.31) .. (305,180) -- (323,180) .. controls (323,183.31) and (325.69,186) .. (329,186) -- (329,233) .. controls (325.69,233) and (323,235.69) .. (323,239) -- (305,239) .. controls (305,235.69) and (302.31,233) .. (299,233) -- cycle ;
\draw  [color={rgb, 255:red, 166; green, 0; blue, 2 }  ,draw opacity=1 ][line width=1.5]  (340,186) .. controls (343.31,186) and (346,183.31) .. (346,180) -- (364,180) .. controls (364,183.31) and (366.69,186) .. (370,186) -- (370,233) .. controls (366.69,233) and (364,235.69) .. (364,239) -- (346,239) .. controls (346,235.69) and (343.31,233) .. (340,233) -- cycle ;
\draw  [color={rgb, 255:red, 126; green, 211; blue, 33 }  ,draw opacity=1 ][line width=1.5]  (379,186) .. controls (382.31,186) and (385,183.31) .. (385,180) -- (403,180) .. controls (403,183.31) and (405.69,186) .. (409,186) -- (409,233) .. controls (405.69,233) and (403,235.69) .. (403,239) -- (385,239) .. controls (385,235.69) and (382.31,233) .. (379,233) -- cycle ;
\draw  [color={rgb, 255:red, 166; green, 0; blue, 2 }  ,draw opacity=1 ][line width=1.5]  (355,178.5) .. controls (355.15,178.5) and (355.27,178.56) .. (355.36,178.68) .. controls (355.46,178.8) and (355.47,178.95) .. (355.42,179.11) .. controls (355.35,179.3) and (355.21,179.45) .. (355,179.56) .. controls (354.76,179.69) and (354.48,179.75) .. (354.17,179.73) .. controls (353.81,179.7) and (353.49,179.59) .. (353.2,179.39) .. controls (352.87,179.16) and (352.64,178.86) .. (352.5,178.5) .. controls (352.35,178.1) and (352.34,177.69) .. (352.47,177.26) .. controls (352.62,176.8) and (352.91,176.39) .. (353.33,176.05) .. controls (353.8,175.67) and (354.35,175.43) .. (355,175.31) .. controls (355.69,175.19) and (356.39,175.23) .. (357.08,175.43) .. controls (357.82,175.65) and (358.45,176.02) .. (358.97,176.55) .. controls (359.52,177.11) and (359.86,177.76) .. (360,178.5) .. controls (360.14,179.28) and (360.05,180.05) .. (359.69,180.8) .. controls (359.32,181.59) and (358.73,182.26) .. (357.92,182.79) .. controls (357.07,183.35) and (356.09,183.69) .. (355,183.81) .. controls (353.86,183.94) and (352.75,183.8) .. (351.67,183.41) .. controls (350.54,183) and (349.61,182.37) .. (348.87,181.51) .. controls (348.09,180.62) and (347.64,179.62) .. (347.5,178.5) .. controls (347.35,177.34) and (347.56,176.22) .. (348.14,175.14) .. controls (348.74,174.01) and (349.64,173.09) .. (350.83,172.37) .. controls (352.07,171.62) and (353.46,171.18) .. (355,171.06) .. controls (356.59,170.94) and (358.12,171.17) .. (359.58,171.75) .. controls (361.09,172.35) and (362.33,173.24) .. (363.3,174.43) .. controls (364.29,175.64) and (364.86,177) .. (365,178.5) .. controls (365,178.5) and (365,178.5) .. (365,178.5) ;
\draw  [color={rgb, 255:red, 0; green, 0; blue, 0 }  ,draw opacity=1 ][fill={rgb, 255:red, 0; green, 0; blue, 0 }  ,fill opacity=1 ] (189,160) -- (269,160) -- (269,260) -- (189,260) -- cycle ;
\draw  [fill={rgb, 255:red, 0; green, 0; blue, 0 }  ,fill opacity=1 ] (164,64.02) -- (174.33,64.02) -- (174.33,60) -- (184,65) -- (174.33,70) -- (174.33,65.98) -- (164,65.98) -- cycle ;
\draw  [fill={rgb, 255:red, 0; green, 0; blue, 0 }  ,fill opacity=1 ] (164,214.02) -- (174.33,214.02) -- (174.33,210) -- (184,215) -- (174.33,220) -- (174.33,215.98) -- (164,215.98) -- cycle ;
\draw  [fill={rgb, 255:red, 0; green, 0; blue, 0 }  ,fill opacity=1 ] (274,64.02) -- (284.33,64.02) -- (284.33,60) -- (294,65) -- (284.33,70) -- (284.33,65.98) -- (274,65.98) -- cycle ;
\draw  [fill={rgb, 255:red, 0; green, 0; blue, 0 }  ,fill opacity=1 ] (274,214.02) -- (284.33,214.02) -- (284.33,210) -- (294,215) -- (284.33,220) -- (284.33,215.98) -- (274,215.98) -- cycle ;

\draw (16,89) node [anchor=north west][inner sep=0.75pt]  [rotate=-270] [align=left] {{\Large User $\displaystyle l$ $ $}};
\draw (16,238) node [anchor=north west][inner sep=0.75pt]  [rotate=-270] [align=left] {{\Large User $\displaystyle p$}};
\draw (50,70) node [anchor=north west][inner sep=0.75pt]  [font=\large] [align=left] {$\displaystyle {\displaystyle N^{l}}$};
\draw (193,13) node [anchor=north west][inner sep=0.75pt]   [align=left] {\textcolor[rgb]{1,1,1}{ \ \ Linear }\\\textcolor[rgb]{1,1,1}{Degrading}\\\textcolor[rgb]{1,1,1}{ \ System}};
\draw (205,81) node [anchor=north west][inner sep=0.75pt]  [font=\Large] [align=left] {$\displaystyle \textcolor[rgb]{1,1,1}{H_{l}( \ )}$};
\draw (90,70) node [anchor=north west][inner sep=0.75pt]  [font=\large] [align=left] {$\displaystyle {\displaystyle N^{l}}$};
\draw (130,70) node [anchor=north west][inner sep=0.75pt]  [font=\large] [align=left] {$\displaystyle {\displaystyle N^{l}}$};
\draw (299,69) node [anchor=north west][inner sep=0.75pt]  [font=\large] [align=left] {$\displaystyle {\displaystyle N^{l}}$};
\draw (379,69) node [anchor=north west][inner sep=0.75pt]  [font=\large] [align=left] {$\displaystyle {\displaystyle N^{l}}$};
\draw (193,163) node [anchor=north west][inner sep=0.75pt]   [align=left] {\textcolor[rgb]{1,1,1}{ \ \ Linear }\\\textcolor[rgb]{1,1,1}{Degrading}\\\textcolor[rgb]{1,1,1}{ \ System}};
\draw (205,231) node [anchor=north west][inner sep=0.75pt]  [font=\Large] [align=left] {$\displaystyle \textcolor[rgb]{1,1,1}{H}\textcolor[rgb]{1,1,1}{_{l}}\textcolor[rgb]{1,1,1}{(}\textcolor[rgb]{1,1,1}{\ }\textcolor[rgb]{1,1,1}{)}$};
\draw (52,213) node [anchor=north west][inner sep=0.75pt]  [font=\large] [align=left] {$\displaystyle {\displaystyle N^{p}}$};
\draw (93,213) node [anchor=north west][inner sep=0.75pt]  [font=\large] [align=left] {$\displaystyle {\displaystyle N^{p}}$};
\draw (132,213) node [anchor=north west][inner sep=0.75pt]  [font=\large] [align=left] {$\displaystyle {\displaystyle N^{p}}$};
\draw (301,212) node [anchor=north west][inner sep=0.75pt]  [font=\large] [align=left] {$\displaystyle {\displaystyle N^{p}}$};
\draw (381,213) node [anchor=north west][inner sep=0.75pt]  [font=\large] [align=left] {$\displaystyle {\displaystyle N^{p}}$};
\draw (339,69) node [anchor=north west][inner sep=0.75pt]  [font=\large] [align=left] {$\displaystyle {\displaystyle A^{l}}$};
\draw (341,213) node [anchor=north west][inner sep=0.75pt]  [font=\large] [align=left] {$\displaystyle {\displaystyle A^{p}}$};

\end{tikzpicture}

\caption{The schema of ABS \cite{kiranyaz2017personalized}: The estimated linear degrading system of the existing user $j$ can be applied to the healthy ECG beats of a new user $p$ in order to synthesize possible abnormal beats for user $p$.}
\label{fig:linear_degrading_system}

\end{figure}

%% file: figures_tikz/linear_transformation_system.tex
\begin{figure}[H]
\captionsetup{font=footnotesize}
\centering

\tikzset{every picture/.style={line width=0.75pt}} 

\begin{tikzpicture}[x=0.65pt,y=0.75pt,yscale=-1,xscale=1]

\draw [color={rgb, 255:red, 65; green, 117; blue, 5 }  ,draw opacity=1 ][line width=1.5]    (6,100) -- (126,99.5) ;
\draw  [color={rgb, 255:red, 65; green, 117; blue, 5 }  ,draw opacity=1 ][line width=1.5]  (26,38.5) -- (41,98.5) -- (11,98.5) -- cycle ;
\draw  [color={rgb, 255:red, 208; green, 2; blue, 27 }  ,draw opacity=1 ][line width=1.5]  (66,38.5) -- (81,98.5) -- (51,98.5) -- cycle ;
\draw  [color={rgb, 255:red, 65; green, 117; blue, 5 }  ,draw opacity=1 ][line width=1.5]  (106,38.5) -- (121,98.5) -- (91,98.5) -- cycle ;
\draw  [color={rgb, 255:red, 0; green, 0; blue, 0 }  ,draw opacity=1 ][fill={rgb, 255:red, 0; green, 0; blue, 0 }  ,fill opacity=1 ] (140,10) -- (230,10) -- (230,110) -- (140,110) -- cycle ;
\draw  [color={rgb, 255:red, 208; green, 2; blue, 27 }  ,draw opacity=1 ][line width=1.5]  (66,38.5) .. controls (66.15,38.5) and (66.27,38.56) .. (66.36,38.68) .. controls (66.46,38.8) and (66.47,38.95) .. (66.42,39.11) .. controls (66.35,39.3) and (66.21,39.45) .. (66,39.56) .. controls (65.76,39.69) and (65.48,39.75) .. (65.17,39.73) .. controls (64.81,39.7) and (64.49,39.59) .. (64.2,39.39) .. controls (63.87,39.16) and (63.64,38.86) .. (63.5,38.5) .. controls (63.35,38.1) and (63.34,37.69) .. (63.47,37.26) .. controls (63.62,36.8) and (63.91,36.39) .. (64.33,36.05) .. controls (64.8,35.67) and (65.35,35.43) .. (66,35.31) .. controls (66.69,35.19) and (67.39,35.23) .. (68.08,35.43) .. controls (68.82,35.65) and (69.45,36.02) .. (69.97,36.55) .. controls (70.52,37.11) and (70.86,37.76) .. (71,38.5) .. controls (71.14,39.28) and (71.05,40.05) .. (70.69,40.8) .. controls (70.32,41.59) and (69.73,42.26) .. (68.92,42.79) .. controls (68.07,43.35) and (67.09,43.69) .. (66,43.81) .. controls (64.86,43.94) and (63.75,43.8) .. (62.67,43.41) .. controls (61.54,43) and (60.61,42.37) .. (59.87,41.51) .. controls (59.09,40.62) and (58.64,39.62) .. (58.5,38.5) .. controls (58.35,37.34) and (58.56,36.22) .. (59.14,35.14) .. controls (59.74,34.01) and (60.64,33.09) .. (61.83,32.37) .. controls (63.07,31.62) and (64.46,31.18) .. (66,31.06) .. controls (67.59,30.94) and (69.12,31.17) .. (70.58,31.75) .. controls (72.09,32.35) and (73.33,33.24) .. (74.3,34.43) .. controls (75.29,35.64) and (75.86,37) .. (76,38.5) .. controls (76,38.5) and (76,38.5) .. (76,38.5) ;
\draw [color={rgb, 255:red, 126; green, 211; blue, 33 }  ,draw opacity=1 ][line width=1.5]    (255,100) -- (375,100) ;
\draw  [color={rgb, 255:red, 126; green, 211; blue, 33 }  ,draw opacity=1 ][line width=1.5]  (261,46) .. controls (264.31,46) and (267,43.31) .. (267,40) -- (285,40) .. controls (285,43.31) and (287.69,46) .. (291,46) -- (291,93) .. controls (287.69,93) and (285,95.69) .. (285,99) -- (267,99) .. controls (267,95.69) and (264.31,93) .. (261,93) -- cycle ;
\draw  [color={rgb, 255:red, 166; green, 0; blue, 2 }  ,draw opacity=1 ][line width=1.5]  (302,46) .. controls (305.31,46) and (308,43.31) .. (308,40) -- (326,40) .. controls (326,43.31) and (328.69,46) .. (332,46) -- (332,93) .. controls (328.69,93) and (326,95.69) .. (326,99) -- (308,99) .. controls (308,95.69) and (305.31,93) .. (302,93) -- cycle ;
\draw  [color={rgb, 255:red, 126; green, 211; blue, 33 }  ,draw opacity=1 ][line width=1.5]  (341,46) .. controls (344.31,46) and (347,43.31) .. (347,40) -- (365,40) .. controls (365,43.31) and (367.69,46) .. (371,46) -- (371,93) .. controls (367.69,93) and (365,95.69) .. (365,99) -- (347,99) .. controls (347,95.69) and (344.31,93) .. (341,93) -- cycle ;
\draw  [color={rgb, 255:red, 166; green, 0; blue, 2 }  ,draw opacity=1 ][line width=1.5]  (316,39.5) .. controls (316.15,39.5) and (316.27,39.56) .. (316.36,39.68) .. controls (316.46,39.8) and (316.47,39.95) .. (316.42,40.11) .. controls (316.35,40.3) and (316.21,40.45) .. (316,40.56) .. controls (315.76,40.69) and (315.48,40.75) .. (315.17,40.73) .. controls (314.81,40.7) and (314.49,40.59) .. (314.2,40.39) .. controls (313.87,40.16) and (313.64,39.86) .. (313.5,39.5) .. controls (313.35,39.1) and (313.34,38.69) .. (313.47,38.26) .. controls (313.62,37.8) and (313.91,37.39) .. (314.33,37.05) .. controls (314.8,36.67) and (315.35,36.43) .. (316,36.31) .. controls (316.69,36.19) and (317.39,36.23) .. (318.08,36.43) .. controls (318.82,36.65) and (319.45,37.02) .. (319.97,37.55) .. controls (320.52,38.11) and (320.86,38.76) .. (321,39.5) .. controls (321.14,40.28) and (321.05,41.05) .. (320.69,41.8) .. controls (320.32,42.59) and (319.73,43.26) .. (318.92,43.79) .. controls (318.07,44.35) and (317.09,44.69) .. (316,44.81) .. controls (314.86,44.94) and (313.75,44.8) .. (312.67,44.41) .. controls (311.54,44) and (310.61,43.37) .. (309.87,42.51) .. controls (309.09,41.62) and (308.64,40.62) .. (308.5,39.5) .. controls (308.35,38.34) and (308.56,37.22) .. (309.14,36.14) .. controls (309.74,35.01) and (310.64,34.09) .. (311.83,33.37) .. controls (313.07,32.62) and (314.46,32.18) .. (316,32.06) .. controls (317.59,31.94) and (319.12,32.17) .. (320.58,32.75) .. controls (322.09,33.35) and (323.33,34.24) .. (324.3,35.43) .. controls (325.29,36.64) and (325.86,38) .. (326,39.5) .. controls (326,39.5) and (326,39.5) .. (326,39.5) ;
\draw  [fill={rgb, 255:red, 0; green, 0; blue, 0 }  ,fill opacity=1 ] (118,64.02) -- (128.33,64.02) -- (128.33,60) -- (138,65) -- (128.33,70) -- (128.33,65.98) -- (118,65.98) -- cycle ;
\draw  [fill={rgb, 255:red, 0; green, 0; blue, 0 }  ,fill opacity=1 ] (233,64.02) -- (243.33,64.02) -- (243.33,60) -- (253,65) -- (243.33,70) -- (243.33,65.98) -- (233,65.98) -- cycle ;

\draw (141,12) node [anchor=north west][inner sep=0.75pt]   [align=left] {\textcolor[rgb]{1,1,1}{{\small  \ \ \ \ \ Linear }}\\\textcolor[rgb]{1,1,1}{{\small Transformation}}\\\textcolor[rgb]{1,1,1}{{\small  \ \ \ \ \ System}}};
\draw (13,78.5) node [anchor=north west][inner sep=0.75pt]  [font=\large] [align=left] {$\displaystyle {\displaystyle N^{l}}$};
\draw (93,78.5) node [anchor=north west][inner sep=0.75pt]  [font=\large] [align=left] {$\displaystyle {\displaystyle N^{l}}$};
\draw (53,78.5) node [anchor=north west][inner sep=0.75pt]  [font=\large] [align=left] {$\displaystyle {\displaystyle A^{l}}$};
\draw (263,72) node [anchor=north west][inner sep=0.75pt]  [font=\large] [align=left] {$\displaystyle {\displaystyle N^{p}}$};
\draw (342,72) node [anchor=north west][inner sep=0.75pt]  [font=\large] [align=left] {$\displaystyle {\displaystyle N^{p}}$};
\draw (305,73) node [anchor=north west][inner sep=0.75pt]  [font=\large] [align=left] {$\displaystyle {\displaystyle A^{p}}$};
\draw (145,80) node [anchor=north west][inner sep=0.75pt]  [font=\Large] [align=left] {$\displaystyle \textcolor[rgb]{1,1,1}{Q_{l}{}_{\rightarrow }{}_{p}( \ )}$};
\draw (32,112) node [anchor=north west][inner sep=0.75pt]   [align=left] {{\Large User $\displaystyle l$ $ $}};
\draw (282,111) node [anchor=north west][inner sep=0.75pt]   [align=left] {{\Large User $\displaystyle p$ $ $}};

\end{tikzpicture}

\caption{Linear Morphology Transformation System.}
\label{fig:linear_transformation_system}

\end{figure}